\documentclass[10pt]{article} 
\usepackage[preprint]{tmlr}

\usepackage{amsmath,amsfonts,bm}

\def\eqref#1{equation~\ref{#1}}

\def\1{\bm{1}}

\DeclareMathAlphabet{\mathsfit}{\encodingdefault}{\sfdefault}{m}{sl}
\SetMathAlphabet{\mathsfit}{bold}{\encodingdefault}{\sfdefault}{bx}{n}

\usepackage{hyperref}
\usepackage{url}
\usepackage{booktabs}       
\usepackage{amsfonts}       
\usepackage{nicefrac}       
\usepackage{microtype}      
\usepackage{xcolor}         
\usepackage{pifont}         
\usepackage{amssymb}        
\usepackage{amsmath}        
\usepackage{graphicx}       
\usepackage{subcaption}     
\usepackage{float}          
\usepackage{tikz}           
\usepackage{svg}            
\usepackage{multirow}       
\usepackage{listings}       

\newfloat{listing}{tbp}{lol}[section]
\floatname{listing}{Listing}

\definecolor{pyblue}{RGB}{38,139,210}
\definecolor{pyteal}{RGB}{42,161,152}
\definecolor{pyred}{RGB}{220,50,47}
\definecolor{pygray}{RGB}{147,161,161}

\newcommand{\cmark}{\textcolor{green}{\ding{51}}}
\newcommand{\xmark}{\textcolor{red}{\ding{55}}}
\newcommand{\wmark}{\textcolor{blue}{$\boldsymbol\sim$}}

\newcommand{\redbg}[1]{\colorbox{red!30}{\texttt{#1}}}
\newcommand{\greenbg}[1]{\colorbox{green!30}{\texttt{#1}}}

\title{TDHook: A Lightweight Framework for Interpretability}

\author{\name Yoann Poupart \email yoann.poupart@lip6.fr \\
      \addr LIP6\\
      Sorbonne University
}

\def\month{MM}  
\def\year{YYYY} 
\def\openreview{\url{https://openreview.net/forum?id=XXXX}} 

\begin{document}

\maketitle

\begin{abstract}
Interpretability of Deep Neural Networks (DNNs) is a growing field driven
by the study of vision and language models.
Yet, some use cases, like image captioning, or domains like Deep Reinforcement Learning (DRL), require complex modelling, with multiple inputs and outputs or use composable and separated networks. As a consequence, they rarely fit natively into the API of popular interpretability frameworks.
We thus present TDHook, an open-source lightweight, generic interpretability framework based on \texttt{tensordict} and applicable to any \texttt{torch} model. 
It focuses on handling complex composed models which can be trained for Computer Vision (CV), Natural Language Processing (NLP), Reinforcement Learning or any other domain.
This library features ready-to-use methods for attribution, probing and a flexible get-set API for interventions, and is aiming to bridge the gap between these method classes to make modern interpretability pipelines more accessible.
TDHook is designed with minimal dependencies, requiring roughly half as much disk space as \texttt{transformer\_lens}, and, in our controlled benchmark, achieves at least a $\times$2 speed-up over \texttt{captum} when running integrated gradients for multi-target pipelines on both CPU and GPU.
In addition, to value our work, we showcase concrete use cases of our library with composed interpretability pipelines in CV and NLP, as well as with complex models in DRL.

\end{abstract}

\section{Introduction}
\label{sec:introduction}

The increasing complexity of Deep Neural Networks (DNNs) has raised significant challenges in understanding their decision-making processes and internal representations. Advancements in large language models \citep{Radford2018ImprovingLU}, computer vision systems \citep{Simonyan2014VeryDC}, and other deep learning architectures have further exacerbated the interpretability challenge by mixing modalities and domains, making it increasingly difficult for researchers and practitioners to debug, let alone trust, these systems.
While the field of explainable AI has grown substantially in recent years, existing efforts in model interpretability predominantly focus on specific domains and architectures. Methods for attribution \citep{Simonyan2013DeepIC, Sundararajan2017AxiomaticAF}, latent manipulation \citep{zou2023representation, Cunningham2023SparseAF}, and weights-based analysis \citep{Dunefsky2024TranscodersFI} have been well-developed, and frameworks like \texttt{captum} \citep{Kokhlikyan2020CaptumAU}, \texttt{zennit} \citep{Anders2021SoftwareFD}, and \texttt{pyvene} \citep{Wu2024pyveneAL} provide specialised tools for these techniques.

However, these solutions can struggle with complex interpretability pipelines, as modern works like concept relevance propagation \citep{Achtibat2022FromAM} or attribution patching \citep{nanda2022attributionpatching} leverage disparate methods. 
They only offer limited compatibility with composed models with multiple inputs or outputs that often require heterogeneous data streams, such as common multi-modal domains \citep{Radford2021LearningTV,Alayrac2022FlamingoAV}, DRL \citep{Liang2017RLlibAF} or other less general domains like finance \citep{Zhou2025FinPRMAD}.
Furthermore, these existing frameworks often lack support for \texttt{tensordict} structures, are limited to specific model architectures, and are not necessarily lightweight or efficient.
These gaps leave practitioners without generic and efficient tools for understanding complex deep learning systems across various domains.

\begin{figure}[t!]
    \centering
    \includegraphics[width=\textwidth]{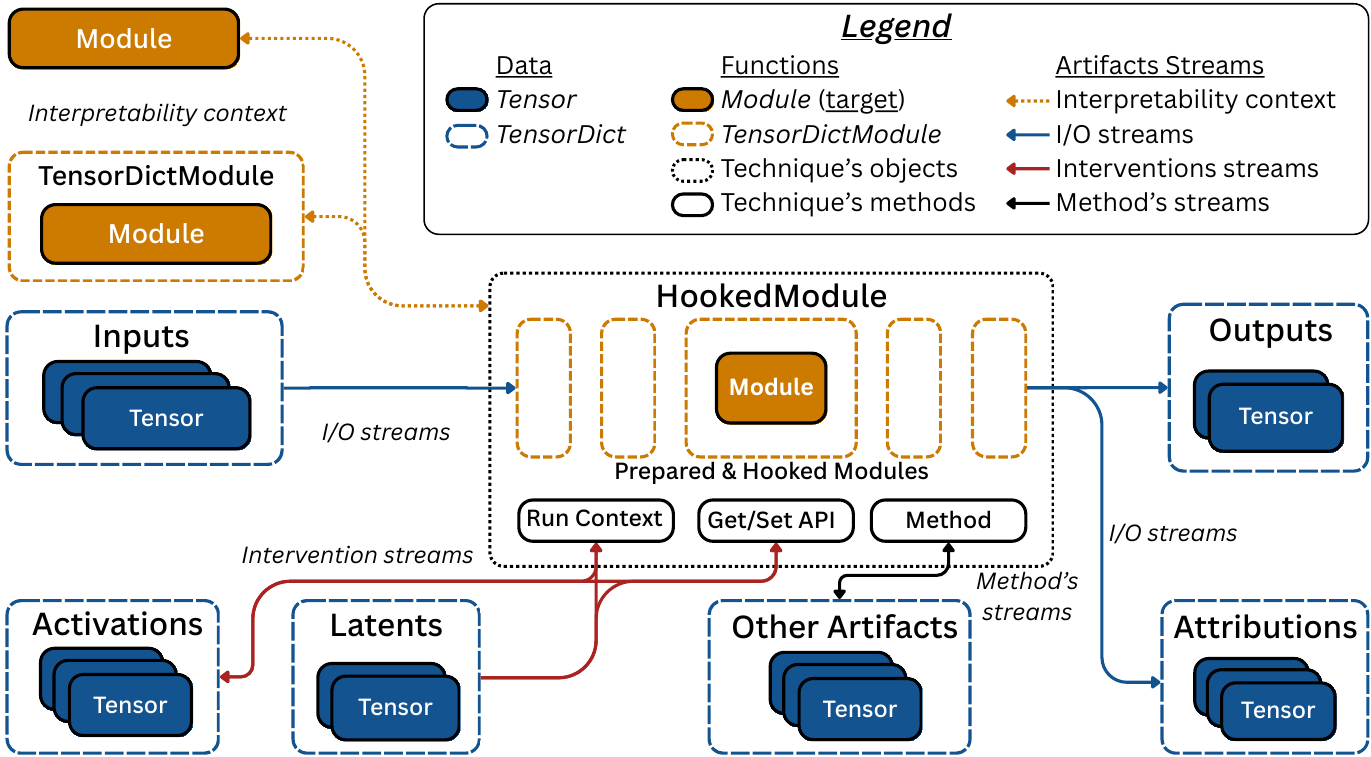}
    \caption{Schematic view of the TDHook framework architecture. The target to interpret is a \texttt{torch} module or a \texttt{TensorDictModule} wrapped in a \texttt{HookedModel}. This object exposes different APIs, like the get-set API described in Section \ref{sec:get_set_api}, while its forward can be modified to fit the needs of different interpretability techniques, see an example in Section \ref{sec:ready_to_use_methods}. The artifacts produced and consumed by the models and methods are materialised as \texttt{TensorDict} objects. More details about the design principles are given in Section \ref{sec:design_principles}, while interactions between the main components of the framework are given in Appendix \ref{app:code}.
    }
    \label{fig:tdhook}
\end{figure}

We thus propose TDHook, a lightweight, generic interpretability framework designed to address these limitations. TDHook implements post-hoc methods that can generate explanations directly from DNNs without constraining their design or needing to extract an interpretable model. The framework is built on two key design principles: (1) compatibility with any \texttt{torch} model through a flexible hooking mechanism, and (2) native support for \texttt{tensordict} structures, which naturally model interpretability by-products like attributions, activations, gradients, and weights. We illustrate the high-level architecture of the framework in Figure \ref{fig:tdhook}, underlining the modularity and broad applicability of the framework.

TDHook makes it easier to compose complex interpretability pipelines across diverse domains, as demonstrated by our use cases study.
Furthermore, it is designed with minimal dependencies, requiring roughly half as much disk space as \texttt{transformer\_lens}, and, in our controlled benchmark, achieves up to a $\times$2 speed-up over \texttt{captum} when running integrated gradients for multi-target pipelines on both CPU and GPU.
We list our contributions as follows:
\begin{itemize}
    \item Introduction of a new interpretability framework,
    TDHook\footnote{The library is provided as supplementary material}, based on \texttt{torch} \citep{Ansel2024PyTorch2F} and \texttt{tensordict} \citep{Bou2023TorchRLAD}.
    \item A set of more than 25 ready-to-use methods for attribution, latent manipulation and weights manipulation, unified through a common interface and only requiring a few lines of code to set up.
    \item A comparative analysis with existing frameworks through stress testing.
    \item Concrete use cases of TDHook applied to language, vision, and DRL models, illustrating multi-step interpretability pipelines.
\end{itemize}

This paper is structured as follows: we present an initial background about interpretability methods and frameworks in Section \ref{sec:background}. Then, we describe the TDHook framework with its design principles in Section \ref{sec:tdhook}, before presenting a comparison study and showcasing concrete use cases in Section \ref{sec:use_cases}. Finally, we discuss the limitations and provide guidance on when to use TDHook in Section \ref{sec:discussion}.

\section{Background}
\label{sec:background}

\subsection{Interpreting Deep Neural Networks}
\label{sec:background_interp}

As we are interested in methods that can be directly applied to a Deep Neural Network (DNN) model, we take 
a post-hoc approach. In Computer Vision (CV) and Natural Language Processing (NLP), many approaches were developed in order to study pre-trained models \citep{Simonyan2014VeryDC, He2015DeepRL, Radford2018ImprovingLU}.
In order to improve code modularity in the TDHook library, we use a categorisation of these methods based on their data stream and produced artifacts.

\paragraph{Attribution methods}
These are typical methods used in CV to understand convolutional networks by visualising important pixels, i.e. saliency maps, \citep{Simonyan2013DeepIC}. 
The key idea is to compute the sensitivity of the output or any intermediate representation with respect to the input or other intermediate representation.
The first developed class of methods were gradient-based \citep{Simonyan2013DeepIC},
with derived methods like integrated gradients \citep{Sundararajan2017AxiomaticAF} or grad-CAM \citep{Selvaraju2016GradCAMVE}, that are mathematically grounded with sensitivity.
Certain methods compute importance maps by occluding part of the input \citep{Zeiler2013VisualizingAU} or, more generally, perturbing the input \citep{Covert2020ExplainingBR}.
Other backpropagation methods like LRP locally define relevance rules, using deep-Taylor expansions, which enable global attribution \citep{Montavon2015ExplainingNC,Bach2015OnPE}. Recent works in NLP focus on the Transformer architecture and its attention mechanism \citep{Vaswani2017AttentionIA}, providing token-level insights \citep{Wiegreffe2019AttentionIN,Achtibat2024AttnLRPAL}.
Finally, some methods go beyond single prediction explanation and aim to provide more conceptual explanations, like
the computation of pre-images from activation maximisation
\citep{Mahendran2015VisualizingDC}, or the finding of maximally activating samples in a dataset \citep{Chen2020ConceptWF}.

\paragraph{Latent manipulation}
This class of methods techniques extends the interpretability of concepts and features by exploring the internal representations learned by models. Latent concepts were introduced in CV with \citep{kim2018interpretability}, 
where concepts are defined as latent space directions, typically computed with linear probes \citep{alain2018understanding,Dreyer2023FromHT}. This field extends the notion of prototypes like perturbed images \citep{Ribeiro2018AnchorsHM}, cropped images \citep{Dreyer2023UnderstandingT} or pre-images \citep{Mahendran2015VisualizingDC}.
These methods were later expanded by the field of representation engineering \citep{zou2023representation}, where such latent features enable locating, editing, erasing or decoding models' knowledge \citep{Meng2022LocatingAE,belrose2023leace, Ghandeharioun2024PatchscopesAU}.
Other works enable causally modifying or analysing the model outputs \citep{rimsky2023steering, Kramar2024AtPAE},
or use sparse autoencoders to elicit interpretable features in language models \citep{Cunningham2023SparseAF}.

\paragraph{Weights-based methods}

These methods focus on interventions and analysis of model parameters themselves, rather than activations or gradients \citep{Saphra2024Mechanistic}. They provide a more granular understanding of model internals by examining pathways and dependencies between models' components, usually neurons or attention heads, and aim to remove the data dependency from
explanations in order to provide out-of-distribution explanations.
Viewing models' components as circuits was first proposed for CNNs \citep{Olah2020ZoomIA} before being formalised for Transformers \citep{elhage2021mathematical}. These circuits revealed peculiar models' components that learned precise mechanisms like induction \citep{Olsson2022IncontextLA}.
Using specific datasets, relevant circuits can be automatically discovered \citep{conmy2023automated},
and can also involve larger models' components at the layer scale \citep{Dunefsky2024TranscodersFI}.
In this framework, operations on the weights become natural, like editing using task vectors \citep{Ilharco2022EditingMW,Hendel2023InContextLC}.
As a consequence, it is also possible to use explanations, from attribution or circuit discovery,
to selectively prune models \citep{Yeom2019PruningBE,Pochinkov2024DissectingLM}.

\subsection{Interpretability Frameworks}
\label{sec:background_frameworks}

\paragraph{Frameworks landscape}
A range of open-source libraries has emerged to operationalise some of the previously described techniques, each with its own focus and design philosophy. \texttt{captum} \citep{Kokhlikyan2020CaptumAU} is the most widely-used \texttt{torch} interpretability library, in terms of downloads and GitHub stars, and concentrates on attribution, providing more than thirty gradient and perturbation-based explainers.
\texttt{zennit} \citep{Anders2021SoftwareFD}, the de facto library for Layer-wise Relevance Propagation methods, offers a unified interface for relevance-based explanations. In the causal-analysis corner, \texttt{pyvene} \citep{Wu2024pyveneAL} provides configuration-based composable interventions while \texttt{nnsight} \citep{FiottoKaufman2024NNsightAN} exposes a general intervention API, based on a getters and setters (get-set API), for altering models' computations at inference time. \texttt{inseq} \citep{Sarti2023InseqAI} specialises in sequence-generation models attribution, most notably by porting the methods from \texttt{captum}. Finally, \texttt{transformer\_lens} \citep{nanda2022transformerlens} targets Transformer models and eases their study by providing a unified access to different model architectures. 

\paragraph{Frameworks comparison}
In Table~\ref{tab:framework_comparison}, we compare these frameworks on features selected to highlight the \texttt{tdhook} library's strengths. 
These features are often missing in the other frameworks and having them all in one place was one of the motivations behind \texttt{tdhook}.
While \texttt{captum} excels in attribution methods, it lacks support for \texttt{TensorDict} and provides no intervention capabilities, limiting its applicability to modern PyTorch workflows and causal analysis. \texttt{nnsight} offers a powerful get-set API but provides no built-in interpretability methods, requiring users to implement techniques from scratch. \texttt{pyvene} supports interventions but lacks \texttt{TensorDict} integration. \texttt{transformer\_lens} is specialised for Transformer architectures only, excluding other model types. \texttt{zennit} focuses solely on LRP methods, while \texttt{inseq} is restricted to sequence generation models.

\texttt{tdhook} combines all these capabilities: it supports arbitrary PyTorch models with \texttt{TensorDict} integration, provides a comprehensive get-set intervention API, offers composable interpretability methods, and includes more than 25 ready-to-use techniques with extensive test coverage.

\begin{table}[t]
\centering
\caption{Comparison of interpretability frameworks on selected features. \textbf{Any Torch}: compatibility with arbitrary PyTorch models; \textbf{TensorDict}: support for \texttt{TensorDict} and \texttt{TensorDictModule}; \textbf{Get-Set}: availability of a generic intervention API; \textbf{Composable}: ability to compose different methods together; \textbf{Methods}: number of ready-to-use interpretability methods; \textbf{Coverage \%}: percentage of the codebase covered by automated tests.}
\label{tab:framework_comparison}
\resizebox{\linewidth}{!}{
\begin{tabular}{l|ccccccc}
\toprule
\textbf{Library} & \textbf{Any Torch} & \textbf{TensorDict} & \textbf{Get-Set} & \textbf{Composable}  & \textbf{Methods} & \textbf{Coverage \%} \\
\midrule
\texttt{captum} & \cmark & \xmark & \xmark & \wmark & 35+ & 93\% \\
\texttt{inseq} & \xmark & \xmark & \xmark & \wmark & 15+ & - \\
\texttt{nnsight} & \cmark & \cmark & \cmark & \xmark & 0 & - \\
\texttt{pyvene} & \cmark & \xmark & \cmark & \cmark & 0 & - \\
\texttt{transformer\_lens} & \xmark & \xmark & \xmark & \xmark & 0 & 75\% \\
\texttt{zennit} & \cmark & \xmark & \xmark & \cmark & 15+ & 99\% \\
\midrule
\textbf{\texttt{tdhook}} & \cmark & \cmark & \cmark & \cmark & 25+ & 95\% \\
\bottomrule
\end{tabular}
}
\end{table}

\section{TDHook}
\label{sec:tdhook}

We now describe the TDHook framework, which is summarised in Figure \ref{fig:tdhook}.

\begin{figure}[t]
    \centering
    \includegraphics[width=0.5\textwidth]{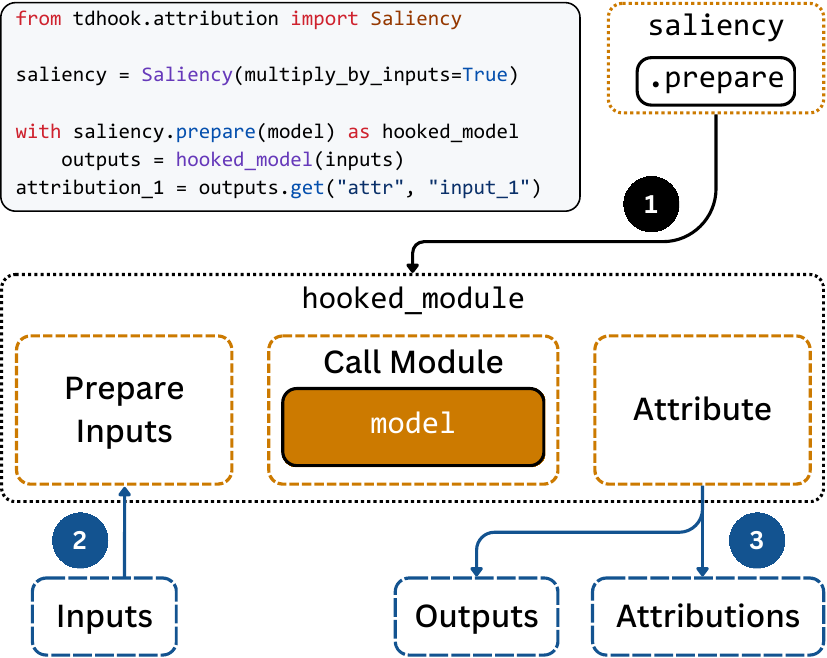}
    \caption{Schematic view of the code execution for the Saliency method in TDHook.
    (1) First, we define a context factory that will prepare a \texttt{Module} or \texttt{TensorDictModule}
    into a \texttt{HookedModel} instance.
    (2) Then, we run this modified model inside the context with \texttt{TensorDict} inputs.
    (3) Finally, we can retrieve the attributions from the output \texttt{TensorDict}.
    }
    \label{fig:tdhook_attr}
\end{figure}

\subsection{Design Principles}
\label{sec:design_principles}

Our objective to create \textbf{composable} interpretability methods in order to handle complex pipelines leads us to consider a \textbf{TensorDict-powered} approach, which provides a strong standard for data and model manipulation. We also aim to provide \textbf{ready-to-use} interpretability methods, making them easy to use even by non-expert users, while keeping the flexibility of a \textbf{generic} framework. Finally, we want our framework to be \textbf{lightweight} for maximal portability.

\paragraph{Composable}
One of the first objectives of TDHook is to be able to implement complex interpretability pipelines, like concept relevance propagation \citep{Achtibat2022FromAM} or attribution patching \citep{nanda2022attributionpatching}. While these methods can be decomposed into simpler methods, providing a simple interface to run them makes them more accessible.
For that, we abstract these methods using a unified approach to how models and data are manipulated, allowing for easy composition. This is made possible by the \texttt{tensordict} library.

\paragraph{TensorDict-powered}
The \texttt{tensordict} library is designed to manage groups of tensors and 
multi-input/output models \citep{Bou2023TorchRLAD}. It provides two primitives: 
\texttt{TensorDict} for carrying data and \texttt{TensorDictModule}
for operations on \texttt{TensorDict} like model predictions.
We focus our efforts around the following postulate:
interpretability methods can be implemented in 
terms of \texttt{TensorDictModule} operating
with \texttt{TensorDict} artifacts.
Indeed, it is clear that interpretability by-products
like activations, gradients, weights or attributions
are collections of tensors, i.e. \texttt{TensorDict},
with different key types (i.e. module names, parameter names, input names, etc.).
And therefore interpretability methods are just \texttt{TensorDictModule},
e.g. attribution methods
take in input a \texttt{TensorDict} (inputs) and return a \texttt{TensorDict} (attributions).

\paragraph{Ready-to-use}
Similarly to \texttt{captum} \citep{Kokhlikyan2020CaptumAU}, we focus our efforts on making interpretability methods easy to use, and provide a wide range of methods,
ready to be used even by non-expert users.
This strength is especially important in order to democratise the interpretability of DDNs.
In that respect, TDHook is not only a generic interpretability framework but also a pragmatic and easy way to integrate interpretability methods. An example using a ready-to-use method is given in Figure \ref{fig:tdhook_attr} and a list of the currently implemented methods is given in Appendix \ref{app:code}.

\begin{figure}[t]
    \centering
    \includegraphics[width=0.5\textwidth]{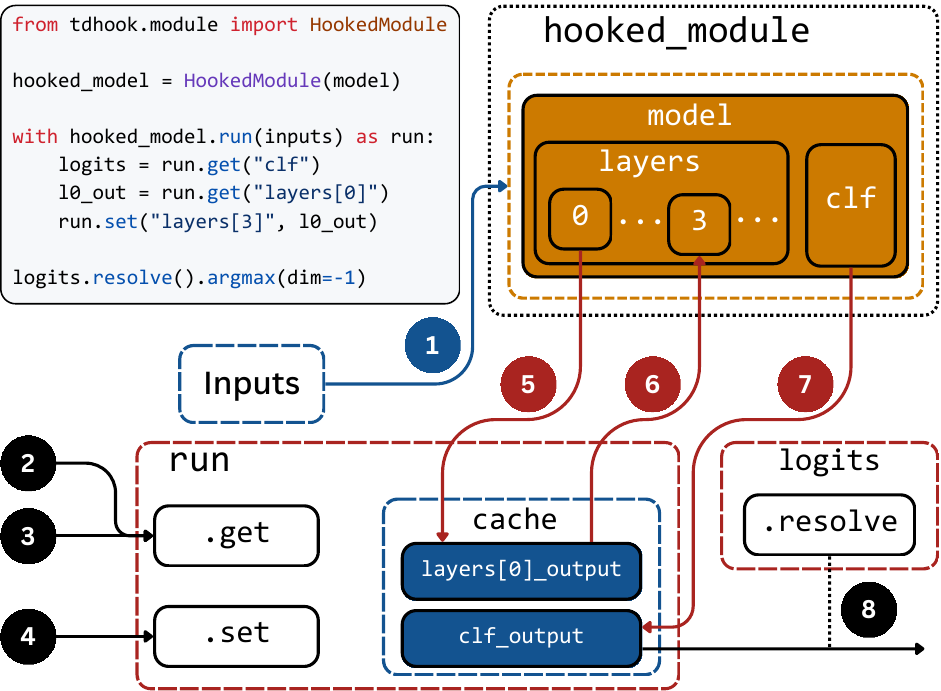}
    \caption{Schematic view of the code execution for an intervention in TDHook.
    (1) After explicitly defining a \texttt{HookedModel} instance, we enter a run context with the given inputs.
    (2-4) We query the run instance to define our intervention scheme
    which registers the required hooks on the model.
    This get-set API returns cache proxies that can be retrieved upon model execution.
    (5-7) At the context exit, we execute the model,  which triggers the registered hooks and populates the cache.
    (8) Finally, we can retrieve any intermediate state by resolving the proxies previously defined.
    }
    \label{fig:tdhook_interv}
\end{figure}

\paragraph{Generic}
TDHook is a generic framework, compatible with any Pytorch model, like \texttt{captum} or \texttt{nnsight}.
It enables users to directly plug their model into the framework and to get started rapidly.
While we mention our design around the \texttt{tensordict} library,
we remain compatible with \texttt{Module} from \texttt{torch} as they can be simply
wrapped with the \texttt{TensorDictModule} class.
Furthermore, in order to enable broader and more customisable methods,
we replicate a flexible intervention API similar to \texttt{nnsight}, based on getters and setters (get-set API), of which we provide an example usage in Figure \ref{fig:tdhook_interv}.

\paragraph{Lightweight}

Some frameworks like \texttt{transformer\_lens}, \texttt{pyvene}, or \texttt{inseq} find their strength in being tailored to work with Transformers, but it comes at a dependency or incompatibility cost.
Yet most of these frameworks rely on \texttt{torch} hooks at their core, which provide the fundamental mechanism for interpretability.
In this respect, TDHook is built around \texttt{torch} hooks in order to be a lightweight interpretability framework with minimal dependencies.
By only admitting two dependencies, \texttt{torch} (already present to run the model) and \texttt{tensordict}, we drastically limit the potential dependency conflicts and the bundle size of the library.
We provide a benchmark of the bundle size of different interpretability libraries in Section \ref{sec:benchmarking}.

\subsection{Ready-to-use Methods}
\label{sec:ready_to_use_methods}

Ready-to-use methods are a powerful adjunct to any library as they enable non-experts to apply interpretability methods. These methods encapsulate common interpretability workflows into simple function calls, abstracting away the complexity of manually setting up hooks, interventions, and data processing pipelines. By providing pre-configured implementations for methods like saliency mapping, latent manipulation, and weights manipulation, users can focus on their research questions rather than implementation details. The underlying complexity of hooking mechanisms, tensor manipulation, and result formatting is handled automatically, making interpretability accessible to researchers with varying levels of technical expertise. This is especially relevant for cross-disciplinary research where DNNs, used as black-box tools, would benefit from being explained.
Figure~\ref{fig:tdhook_attr} illustrates the execution flow for a saliency method, showing how the context factory prepares the model and how attributions are retrieved. For comprehensive implementation details and code examples, we refer the reader to Appendix~\ref{app:code} and the supplementary materials.

\subsection{General Get-Set API}
\label{sec:get_set_api}

As already proposed by \citep{FiottoKaufman2024NNsightAN}, the get-set API is a versatile way to interact with a model. This pattern separates the definition of interventions (with getters and setters) from their execution and result retrieval. 
We adopt a more explicit approach using \texttt{HookedModel} methods that provide fine-grained control over model interventions and state retrieval.
The get-set API allows researchers to first define their intervention scheme, specifying which layers to hook, what modifications to apply, and which intermediate states to capture, before executing the model. This separation enables complex intervention strategies to be built incrementally and ensures that all interventions are applied consistently during a single forward pass and a potential backward pass. 
The API returns cache proxies that act as placeholders for future values, which are then populated during execution and can be resolved afterwards to retrieve the actual intermediate representations.
The model can also be executed in a run context offering the same API, triggering the registered hooks and populating the cache with the intermediate representations. This context automatically handles hooks deregistration, temporary cache configuration and offers other capabilities like stopping the computation at any layer.
Figure~\ref{fig:tdhook_interv} demonstrates the intervention workflow, showing how hooks are registered during the setup phase and how intermediate states are retrieved after execution. Implementation details can be found in Appendix~\ref{app:code} and in the supplementary materials.

\section{Benchmarking}
\label{sec:benchmarking}

In order to compare our library beyond the accuracy of the implemented methods, we introduce two benchmarks to position our library in the framework landscape.

\paragraph{Performance benchmark} 
In order to fairly compare the different libraries, we use the tasks which are showcased as a "Get Started" task in their respective documentation, and we reproduce them in TDHook, checking the consistency of the results\footnote{Caching in \texttt{transformer\_lens} is not reproduced exactly due to the model modifications needed.}. For each task, we run multiple experiments with different seeds, model sizes, batch sizes and other parameters, consisting of more than a hundred variations. Our experiments include models ranging from small MLPs to transformers (using the GPT-2 family), with up to 1B parameters. The experiments were conducted with batch sizes up to 100k for MLPs, in order to explore the limits of common use cases like activation caching. For reproducibility purposes, we give more details in Appendix \ref{app:benchmarking} and provide the complete code in the supplementary material. We also depict the exact hardware configuration used for the benchmark in Appendix \ref{app:benchmarking}.
In Figure \ref{fig:performance_benchmark}, we report the aggregated results of our benchmark. We measure the relative performance of each library against TDHook across different metrics (time, RAM, VRAM) and different setups (CPU only or CPU+GPU).
Each task run is performed in isolation and always starts with a "spawn" (model setup or other preparations without computation) for which the metrics are also reported. This kind of benchmark is useful to stress-test the different libraries and to highlight the different trade-offs between them in terms of resource consumption. For example, while Captum is optimised in terms of memory usage, it is slower due to its gradient computation strategy\footnote{Captum only computes the gradient with respect to scalar values instead of vector values.}, it does not scale well with the batch size. TransformerLens has a setup overhead in terms of memory usage in order to convert the model to a unified format; it does not scale well with the model size.

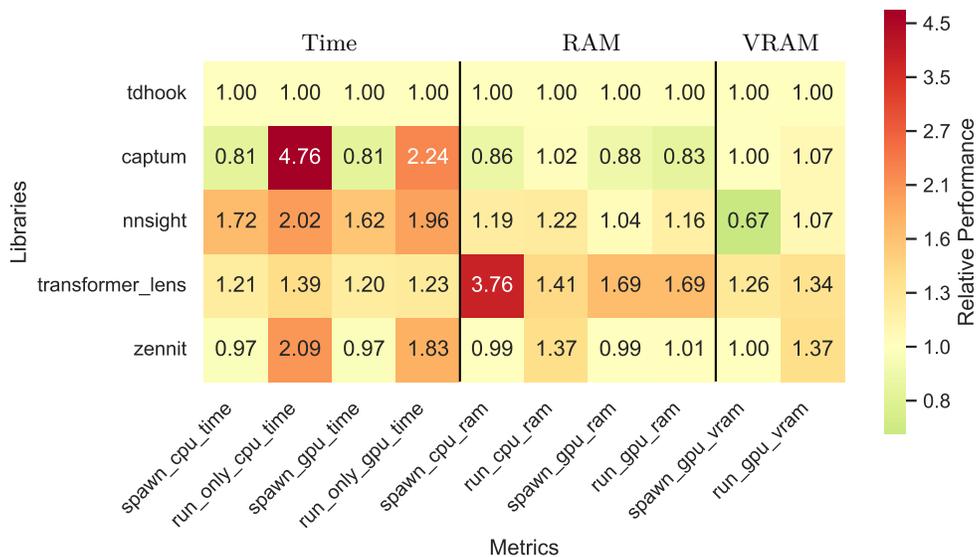
\begin{figure}[t]
    \centering
    \begin{tikzpicture}
        \node[anchor=south west,inner sep=0] (image) at (0,0) {
            \includegraphics[width=0.8\textwidth]{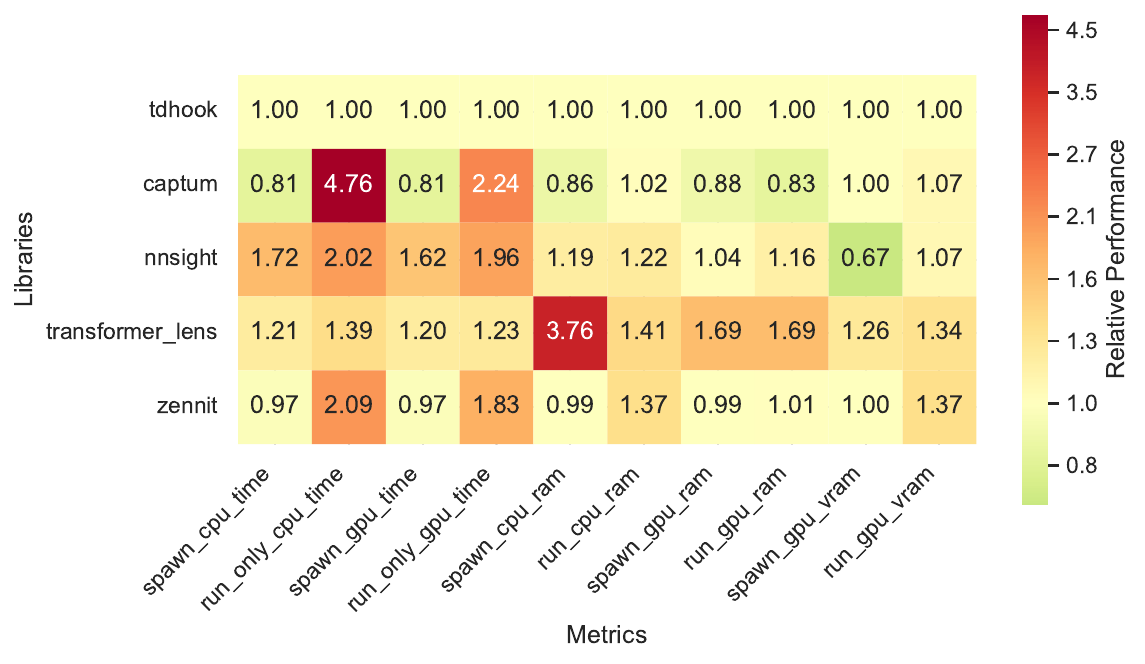}
        };
        \node[anchor=south west,inner sep=0] at (0.245\textwidth,0.42\textwidth) {\small Time};
        \node[anchor=south west,inner sep=0] at (0.455\textwidth,0.42\textwidth) {\small RAM};
        \node[anchor=south west,inner sep=0] at (0.6\textwidth,0.42\textwidth) {\small VRAM};

        \draw[black, thick] (0.373\textwidth,0.154\textwidth) -- (0.373\textwidth,0.412\textwidth);
        \draw[black, thick] (0.579\textwidth,0.154\textwidth) -- (0.579\textwidth,0.412\textwidth);
    \end{tikzpicture}
    \caption{Relative performance of interpretability frameworks on their "Get Started" task compared against tdhook implementation (lower is better). The colours follow a logarithmic scale centred on the tdhook relative performance. More technical details and results are given in Appendix \ref{app:benchmarking}, and the individual task results are provided in supplementary material.}
    \label{fig:performance_benchmark}
\end{figure}

\begin{figure}[thbp]
    \centering
    \subfloat[Memory size]{
        \includegraphics[width=0.45\textwidth]{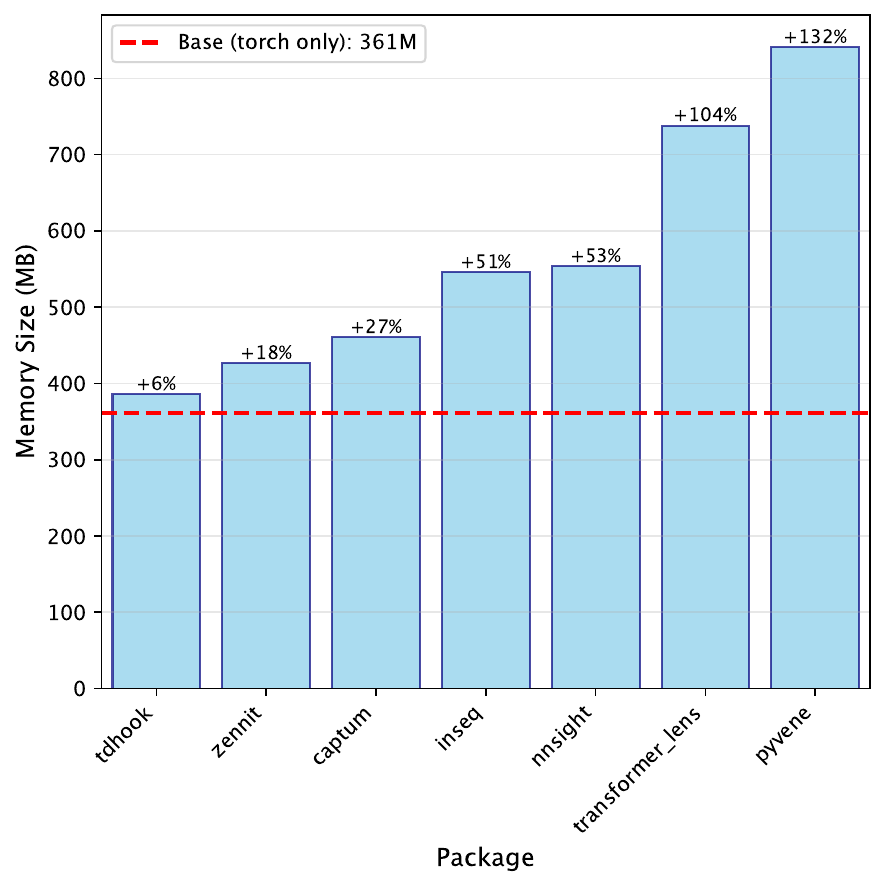}
    \label{fig:memory_sizes}
    }
    \hfill
    \subfloat[Inode count]{
        \includegraphics[width=0.45\textwidth]{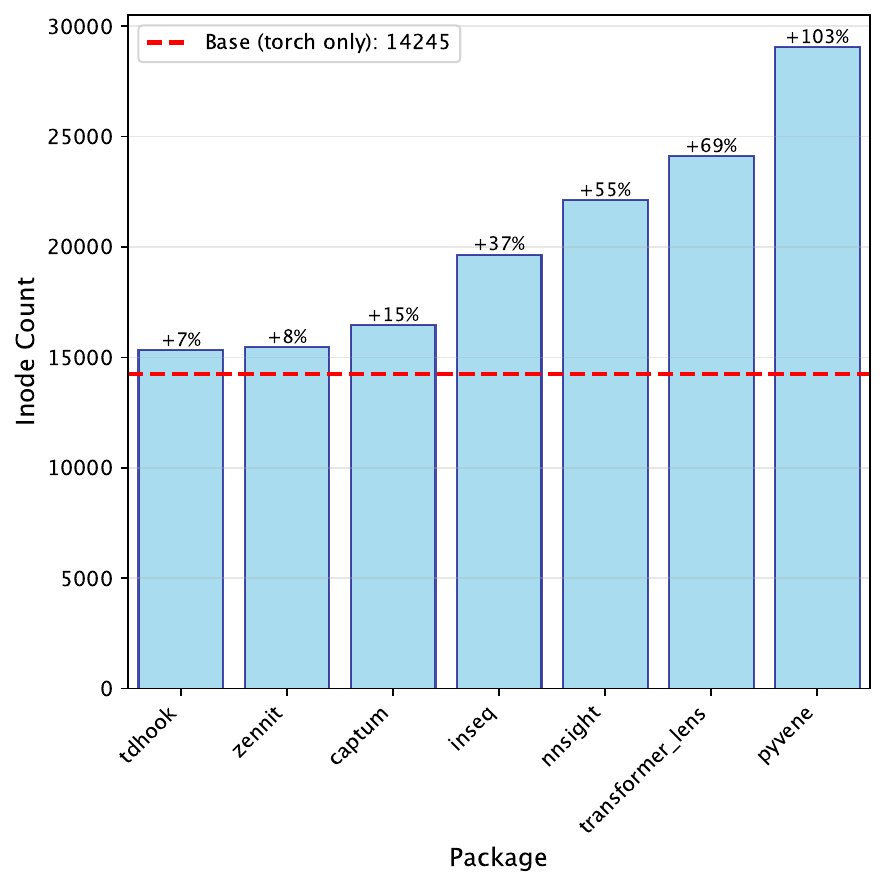}
        \label{fig:inode_counts}
    }
    \caption{Bundle size comparison across different interpretability frameworks by measuring the memory size (a) and the inode count (b).
    For each measurement, we compare with only installing \texttt{torch}; this baseline space could also be reduced by choosing a lighter version of \texttt{torch}, e.g. \texttt{torch-cpu}.}
    \label{fig:bundle_size_benchmark}
\end{figure}

\paragraph{Bundle size benchmark}
The second benchmark compares the bundle size of different interpretability frameworks. We considered two different metrics, which are synonyms of light-weight disk usage: the memory space and the number of inodes. While the disk space is commonly used to partition the disk among users, the number of inodes can also be limited in certain scenarios, like HPC clusters. The results in Figure \ref{fig:bundle_size_benchmark} differentiate libraries that focus on being model agnostic versus libraries that focus on models from the Transformers library \citep{Wolf2020TransformersSO}. 
These measurements are conducted in the independent virtual environments obtained by installing each framework individually. We managed those dependencies by using \texttt{uv} \citep{uv}; more details about the benchmarks are given in Appendix \ref{app:benchmarking}.

\section{Use Cases}
\label{sec:use_cases}

We now showcase examples demonstrating specific features that motivated the design of TDHook, and extend results from previous works. These examples are spread across natural language processing, computer vision and reinforcement learning.

\subsection{Complex Pipelines}

Modern interpretability pipelines have evolved into complex multi-step procedures, mixing different methods in order to derive more comprehensive explanations. We here focus on concept attribution and attribution patching, which are described in the following paragraphs.

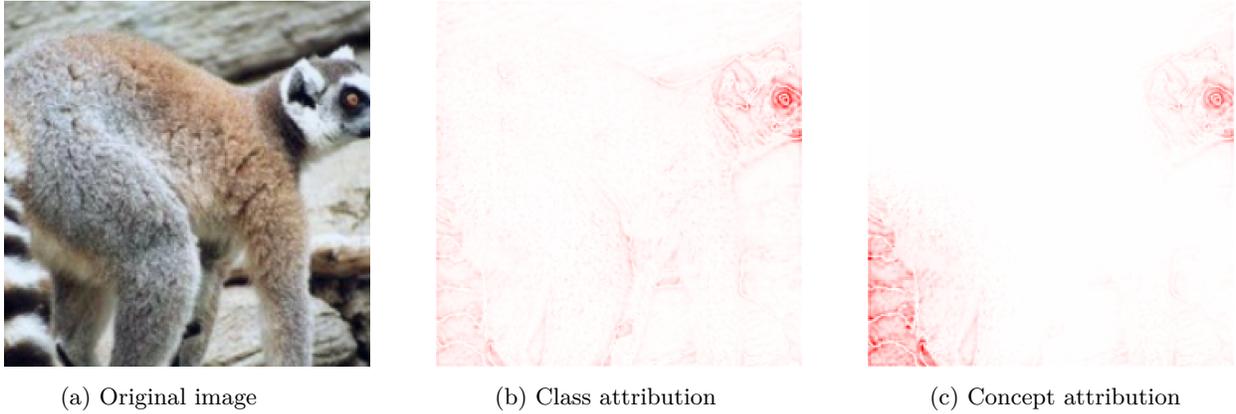
\begin{figure}[tb]
    \centering
    \begin{tikzpicture}
        \node[anchor=south west,inner sep=0] (image) at (0,0) {
            \includegraphics[width=\textwidth]{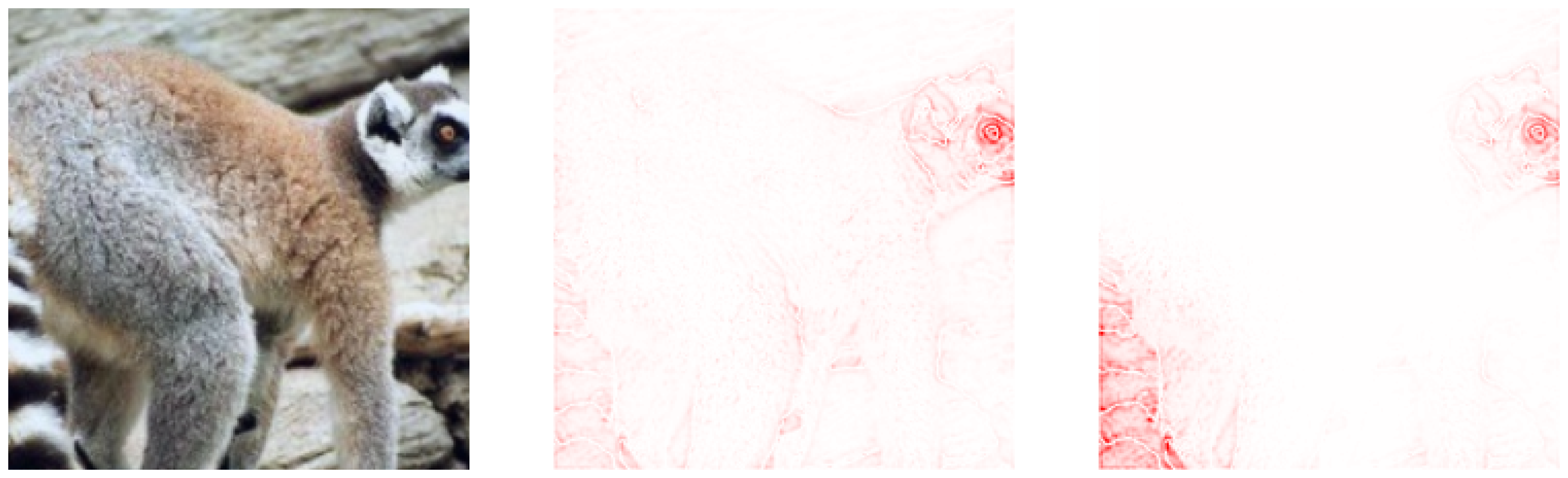}
        };
        \node[anchor=south west,inner sep=0] at (0.05\textwidth,-0.03\textwidth) {\small (a) Original image};
        \node[anchor=south west,inner sep=0] at (0.4\textwidth,-0.03\textwidth) {\small (b) Class attribution};
        \node[anchor=south west,inner sep=0] at (0.75\textwidth,-0.03\textwidth) {\small (c) Concept attribution};
    \end{tikzpicture}
    \caption{Concept attribution with the LRP method. (a) The original image being explained. (b) The class attribution heatmap. (c) Attribution of the channel associated with the concept "striped" selected by activation maximisation.}
    \label{fig:concept_attribution_lrp}
\end{figure}

\paragraph{Concept attribution}
Contrary to traditional attribution methods that focus on explaining the output of the model, e.g. the class predicted, with respect to the input, concept attribution methods focus on explaining the output with respect to specific concepts, illustrated in Figure \ref{fig:concept_attribution_lrp}. Most of these methods rely on conditionally propagating the attribution with respect to specific channels, like conductance for integrated gradients \citep{Dhamdhere2018HowII} or concept relevance propagation for LRP \citep{Achtibat2022FromAM}. These channel concepts are usually discovered or visualised using activation maximisation \citep{Chen2020ConceptWF} or relevance maximisation \citep{Achtibat2022FromAM}. But since concept activation vectors (CAVs) generalise these methods, we introduce a propagation method based on the TCAV scores \citep{kim2018interpretability}. Different methods for producing CAVs were already compared in \citep{Dreyer2023FromHT}, but not in the context of concept attribution.

\begin{figure}[tb]
    \centering
    \includegraphics[width=\textwidth]{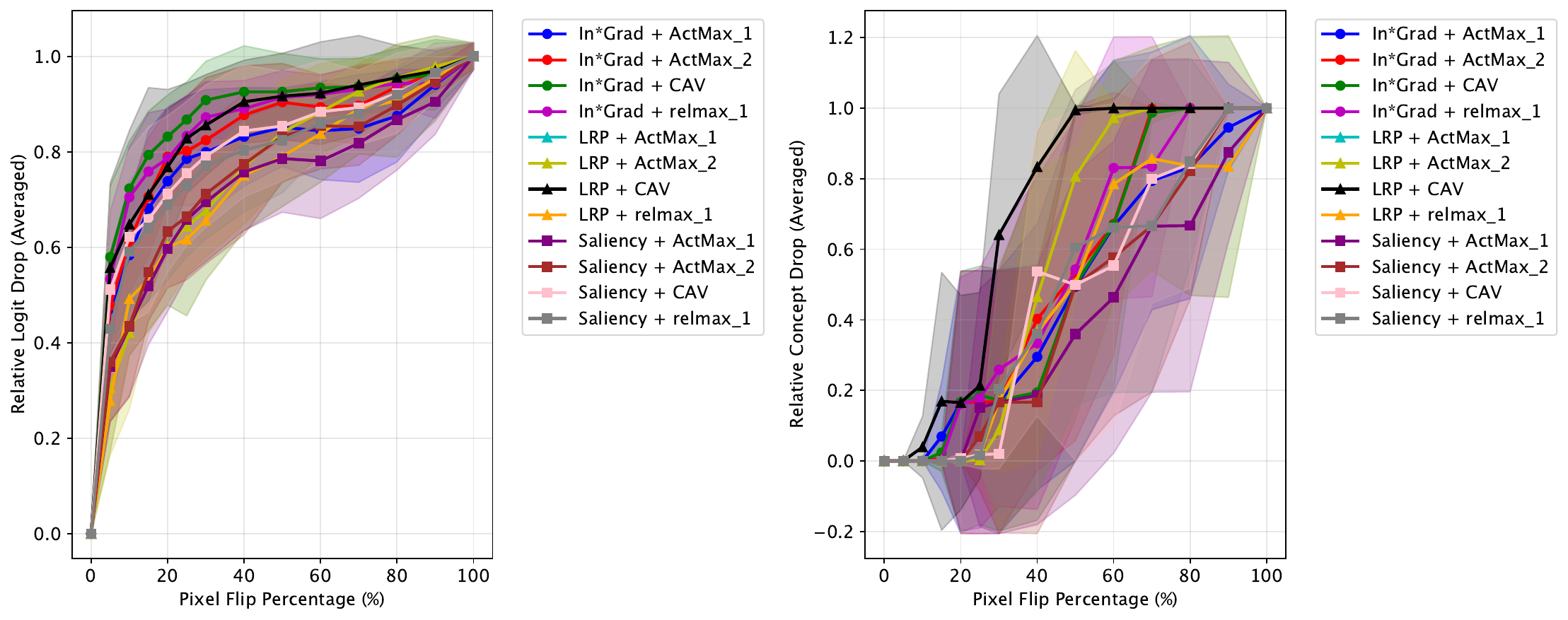}
    \caption{Concept attribution evaluation for the concept "striped". We report the change in the class logit and the concept probability when the input is randomised according to the relevance scores. The ablated pixels are replaced by the mean of the input image.}
    \label{fig:concept_attribution_evaluation}
\end{figure}

Concept attribution is complex, as it is a composed method, and it is important to be able to compare across a variety of possible combinations of methods. As all methods are implemented in TDHook, it is possible to easily compare them by simply switching the method used, enabling users to chose their prefered notion of concept or attribution. For our experiments, we use the VGG16 model \citep{Simonyan2014VeryDC}, and extracted concepts like "striped" using the texture dataset from \citep{Cimpoi2013DescribingTI}.
In order to evaluate the methods, we randomise the input according to the relevance scores, i.e. pixel flipping \citep{Bach2015OnPE}, and record the change in the class logit and the concept probability. The results, reported in Figure \ref{fig:concept_attribution_evaluation}, 
show that using the TCAV scores gives attribution that is more faithful to the concept. TCAV marginally outperforms the other methods according to pixel-flipping evaluation, as shown in Figure \ref{fig:concept_attribution_evaluation}. Yet, qualitative analysis suggests that the obtained attribution is not localised, e.g., showing relevance spread beyond the lemur's tail. Therefore other metrics should be used to evaluate the quality of the attribution, like localisation or robustness metrics \citep{Hedstrm2022QuantusAE}.

\begin{figure}[thbp]
    \centering
    \includegraphics[width=0.8\textwidth]{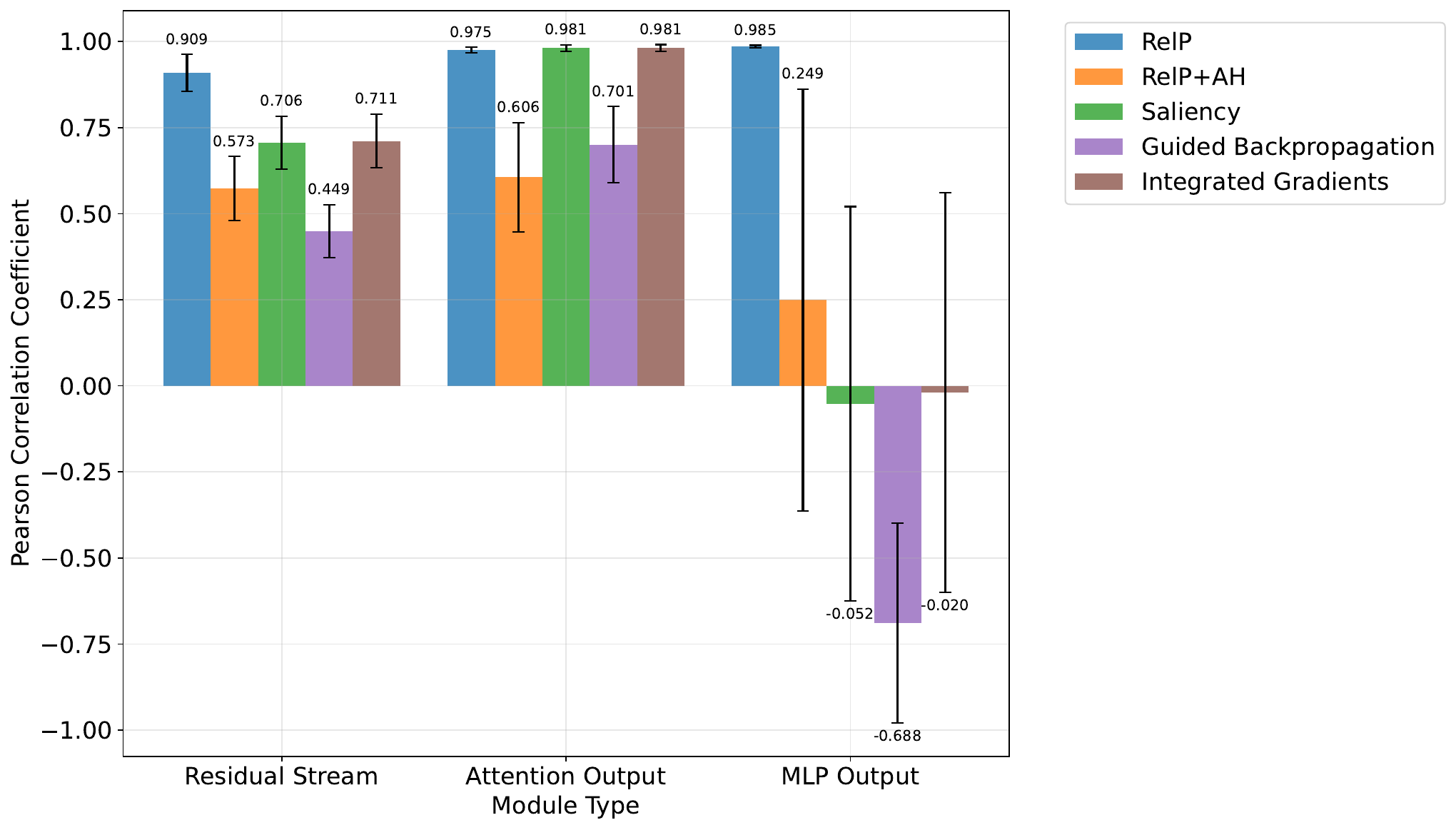}
    \caption{Pearson correlation coefficients between different attribution patching methods and the activation patching method. The values are averaged for different models (GPT-2 family), and the error bars represent the standard deviation. This metric does not account for amplitude differences between the methods.
    }
    \label{fig:attribution_patching_evaluation}
\end{figure}

\paragraph{Attribution patching}
Activation patching, also known as causal mediation analysis \citep{Vig2020InvestigatingGB}, is a method to understand the contribution of different components of the model to the output. It provides a powerful causal analysis but is costly to compute, as it requires as many forward passes as there are components to patch.
Attribution patching \citep{nanda2022attributionpatching} was proposed as an alternative using a first-order approximation, only requiring two forward passes and one backward pass. Recent work has replaced the gradient by a relevance score obtained using LRP, i.e. RelP \citep{Jafari2025RelPFA}. We explore the generalisability of this idea by replacing gradients by latent relevances. And show that the results of the RelP method are sensitive to the LRP rules used, notably the results using the AH-rule, introduced in the paper {Jafari2025RelPFA} but not used for the experiments, are presented in Appendix \ref{app:use_cases_details}. The results in Figure \ref{fig:attribution_patching_evaluation} show that only methods closely related to the gradient are relevant for attribution patching.


\subsection{Complex Models}

In addition to the pipelines, models can also expose a complex structure, with multiple inputs and outputs. We here focus on multi-output models, which are common in DRL.

\paragraph{Analyse multi-output models} It is common for models in RL to have multiple inputs and outputs. We here take the example of a chess model which is trained with AlphaZero \citep{Silver2018AGR}. In particular, we choose a network from \citet{McIlroyYoung2020AligningSA}, which outputs the policy over the legal moves and the win-draw-lose probabilities for the current position. More details about the models used are given in Appendix \ref{app:use_cases_details}. 
In Figure \ref{fig:saliency} we show saliency maps for the policy and the value heads obtained using gradient attribution \citep{Simonyan2013DeepIC}. The position chosen, taken from \citet{lichess_puzzles}, is a mate in 2 for the black player (\texttt{e3e1}, \texttt{b1a2}, \texttt{e1a1}).
Attributing the best move gives a positive saliency to the rook and the bishop, while the adversary king has a negative saliency. This reflects the move sequence leading to the mate: the rook checks the king, the king flees to the side, and the rook mates while being defended by the bishop.
For the win probability attribution, the attacking pieces have a high positive saliency, while the black king and the pawn have a negative saliency, as they are weak points.
The adversary queen also has a high positive saliency, as it can be captured by the knight.
While the predictive power of the saliency is limited, it can still provide basic insights about the model.

\begin{figure}[tb]
    \centering
    \begin{subfigure}[b]{0.37\textwidth} 
        \centering
        \includegraphics[height=0.2\textheight]{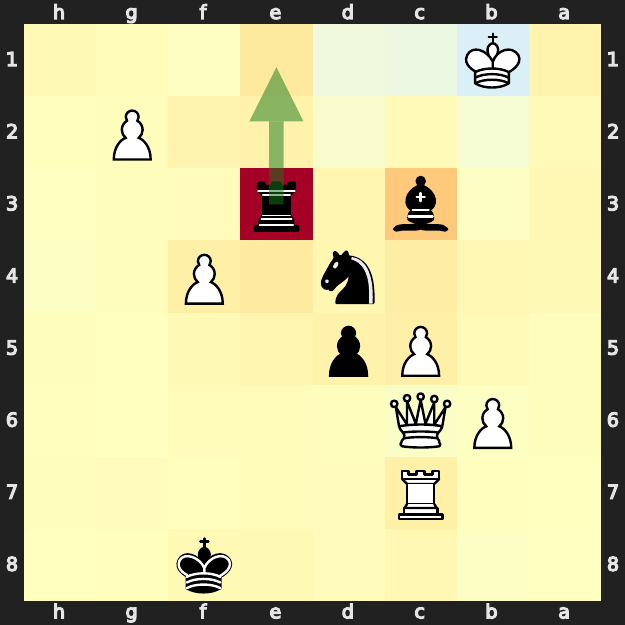}
        \caption{Best move saliency}
        \label{fig:best_move_saliency_board}
    \end{subfigure}
    \begin{subfigure}[b]{0.1\textwidth}
        \centering
        \includegraphics[height=0.25\textheight]{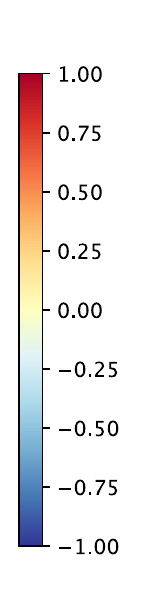}
    \end{subfigure}
    \begin{subfigure}[b]{0.37\textwidth} 
        \centering
        \includegraphics[height=0.2\textheight]{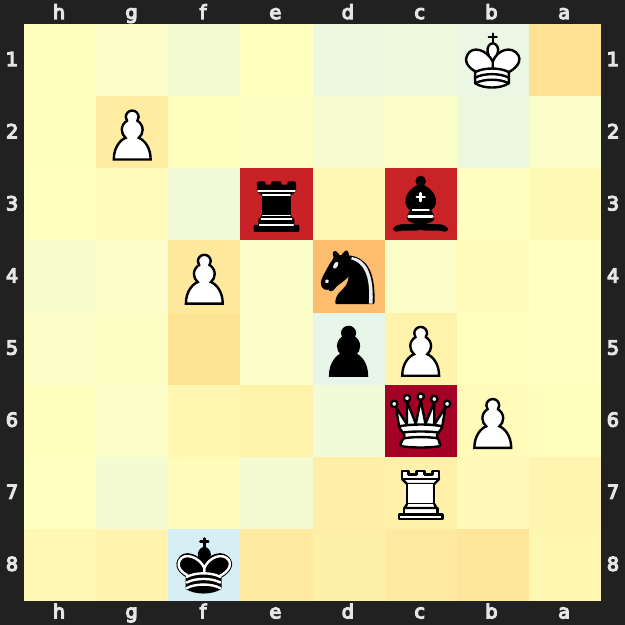}
        \caption{Win saliency}
        \label{fig:win_saliency_board}
    \end{subfigure}
    
    \caption{A same board position analysed through the prism of the policy head and the value head using gradients. The attributions are scaled by the absolute maximum value such that the colour meaning is shared for (a) and (b). For policy attribution (a), the output explained is the logit of the best legal move, which is plotted as a green arrow, and for the value attribution (b), the output explained is win probability. 
    }
    \label{fig:saliency}
\end{figure}

\begin{figure}[tb]
    \centering
    \includegraphics[width=0.8\textwidth]{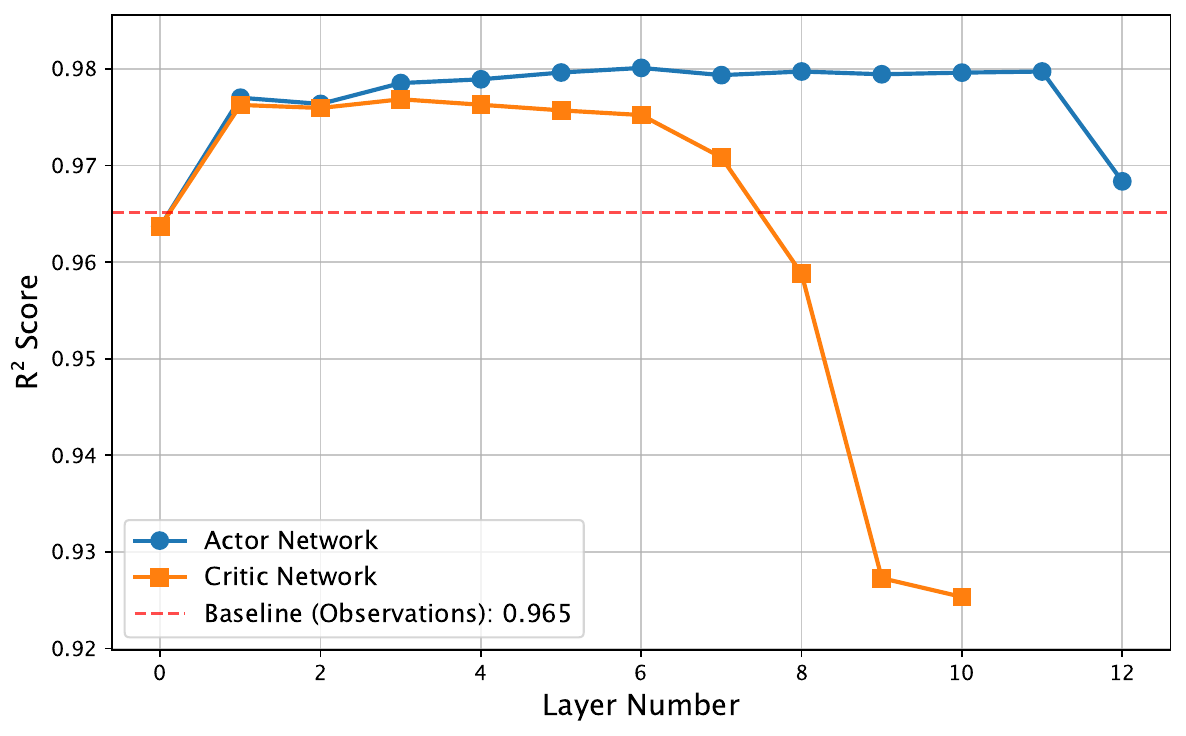}
    \caption{Action probing for the policy and value networks of a PPO agent trained on the inverted double pendulum environment. 
    We report the $R^2$ scores between the predicted action and the model's action for relevant layers evaluated on a different set of activations. We compare with a baseline trained from the raw observations.}
    \label{fig:action_probing}
\end{figure}

\paragraph{Analyse composed models}
While the previous example showcased a shared backbone for the policy and the value, it is not always the case.
Especially in decentralised multi-agent settings, the value is often independent from the policy, and each agent can have its own policy and value. 
Therefore, it means that you have multiple networks to explain. 
Yet, in TorchRL, these models are grouped in a single \texttt{TensorDictModule}, the loss, which can be directly wrapped in TDHook.
For our illustration, we train an agent with PPO \citep{Schulman2017ProximalPO} on the inverted double pendulum environment \citep{Todorov2012MuJoCoAP}. More details about the agent studied are given in Appendix \ref{app:use_cases_details}.
In this case, we decided to explore how the predicted action
is represented throughout the internal states of this agent
using linear probing \citep{alain2018understanding}.
Figure \ref{fig:action_probing} shows the action probing in the agent, performed both on the policy and the value by wrapping the loss module.
For each layer (pre- and post-activation), a linear regression probe is trained to predict the model's action.
We can see that the policy head contains more information about the action than the value head, as expected, but the value head still overperforms the baseline trained from the raw observations. 
For the policy, the last layers contain less information 
about the predicted action because it is highly compressed, only predicting the mean and standard deviation of a Tanh-Normal distribution. For the value, the information about the action is not trackable in the last layers, and the probe overfitted on the training data, as it can be expected when the signal weakens.

\section{Discussion}
\label{sec:discussion}

We now discuss the positioning of TDHook in the broad framework landscape with regard to its relevance to specific use cases, and propose avenues for future work.

\subsection{When to Use TDHook}

TDHook is most beneficial in scenarios where researchers need to analyse models that produce multiple outputs or are composed of several sub-modules, a situation frequently encountered in reinforcement-learning pipelines. Because the library natively embraces \texttt{TensorDict} and \texttt{TensorDictModule}, it can ingest the heterogeneous tensors emitted by these models without any additional data-wrangling layer. Its generic get-set API also makes TDHook a compelling choice for rapid prototyping: a new attribution or intervention can often be implemented in a handful of lines while retaining compatibility with the core abstractions. It is also relevant to build composed interpretability pipelines based on multiple ready-to-use methods which can be chained together.
Finally, our benchmarks in Section~\ref{sec:benchmarking} demonstrate that TDHook maintains a small installation footprint and competitive runtime memory usage; this makes it a sensible default on resource-constrained hardware such as edge devices running torch-CPU.

\subsection{When to Use Other Frameworks}

Despite this versatility, other interpretability frameworks can be a better fit in specific circumstances. \texttt{captum} remains unrivalled when an analyst requires an extensive catalogue of ready-made attribution algorithms and is not concerned with multi-input or multi-output settings. Transformer-centric libraries such as \texttt{transformer\_lens} or \texttt{inseq} expose specialised utilities and visualisations that accelerate studies of language models at the expense of broader architectural support. Configuration-driven causal experimentation may be easier to conduct with \texttt{pyvene}, which provides configuration-based recipes for large-scale sweeps, while \texttt{zennit} offers broader implementations for LRP and its variants. Finally, \texttt{nnsight} bundles interpretability routines within a neat intervention API while enabling experiments to be run on models hosted on the NDIF servers.

\subsection{Future Work}

Future releases of TDHook will focus on three complementary directions. First, we will broaden the set of integrated methods by porting additional attribution, probing and causal-manipulation techniques, with an emphasis on more complex pipelines. This includes extending our work to more sophisticated models, like transformers with chain-of-thought modules, requiring large-scale analyses. Second, we plan to exploit the advanced features of \texttt{tensordict}, in particular memory-mapped tensors and zero-copy sharing, to further reduce the peak RAM footprint during large-scale analyses. Lastly, we aim to extend support to distributed and heterogeneous execution environments so that TDHook can seamlessly operate in multi-GPU and multi-node training loops. These technical improvements will be accompanied by richer documentation and end-to-end tutorials that showcase the library on vision, language and control tasks, with primary focus on composed interpretability pipelines.

\subsubsection*{Broader Impact Statement}
As an interpretability tool, the framework proposed does
not raise immediate ethical risks beyond those inherent to interpretability research.
However, the goal of making interpretability methods easier to use and more accessible to non-experts
can exacerbate misuse cases.
The most concerning issues are:
\begin{itemize}
    \item Creating misleading explanations if results are over-interpreted without rigour or without sufficient domain expertise.
    \item Model control abuse, e.g. using interpretability methods to bypass safety measures or to create harmful content.
\end{itemize}
Therefore, we encourage users to use the framework with caution and to be aware of the limitations of interpretability methods.


\bibliography{position,tdhook,custom}
\bibliographystyle{tmlr}

\appendix
\section{Benchmarking Details}
\label{app:benchmarking}

We present here more detailed results about the performance benchmarks presented in Section \ref{sec:benchmarking}.

\paragraph{Performance benchmark}

\begin{figure}[thbp]
    \centering
    \subfloat[CPU spawn time]{
        \includegraphics[width=0.45\textwidth]{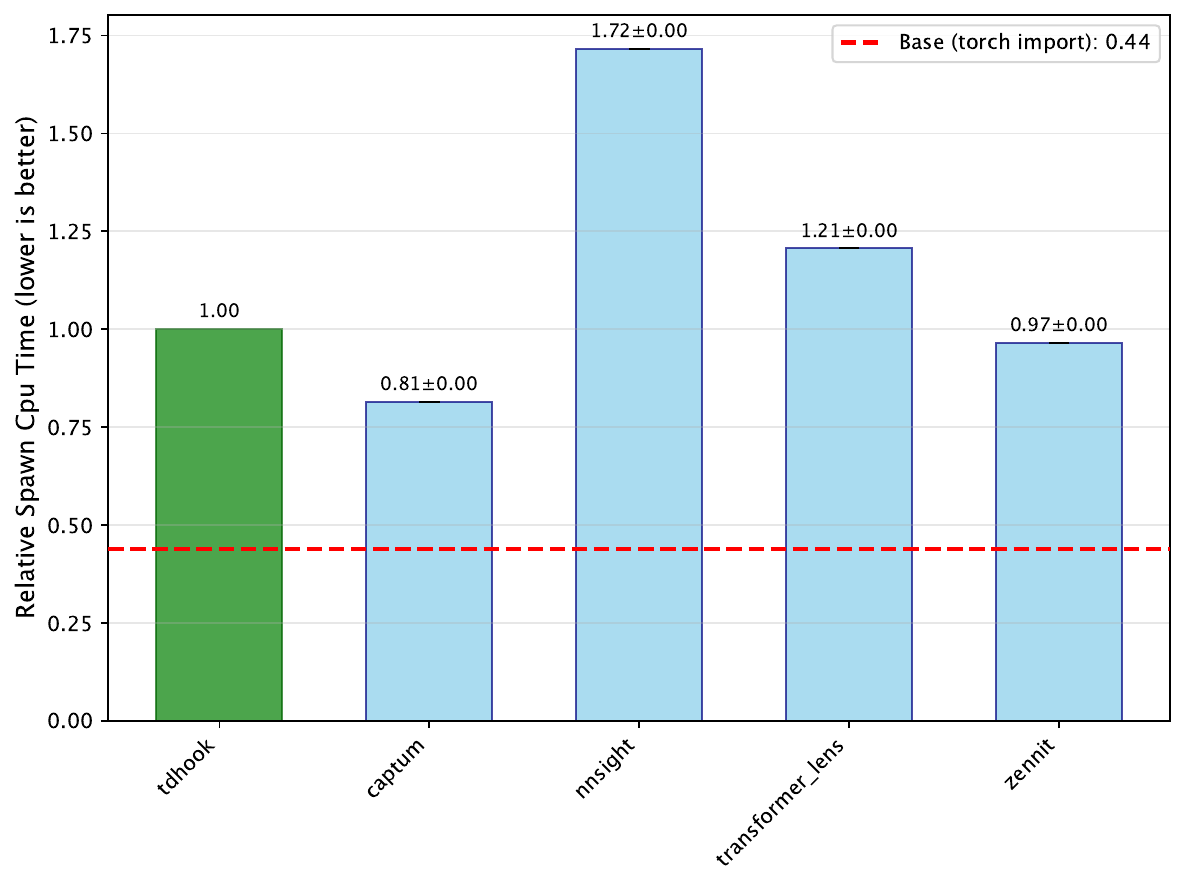}
        \label{fig:benchmarking_spawn_cpu_time}
    }
    \hfill
    \subfloat[CPU run time]{
        \includegraphics[width=0.45\textwidth]{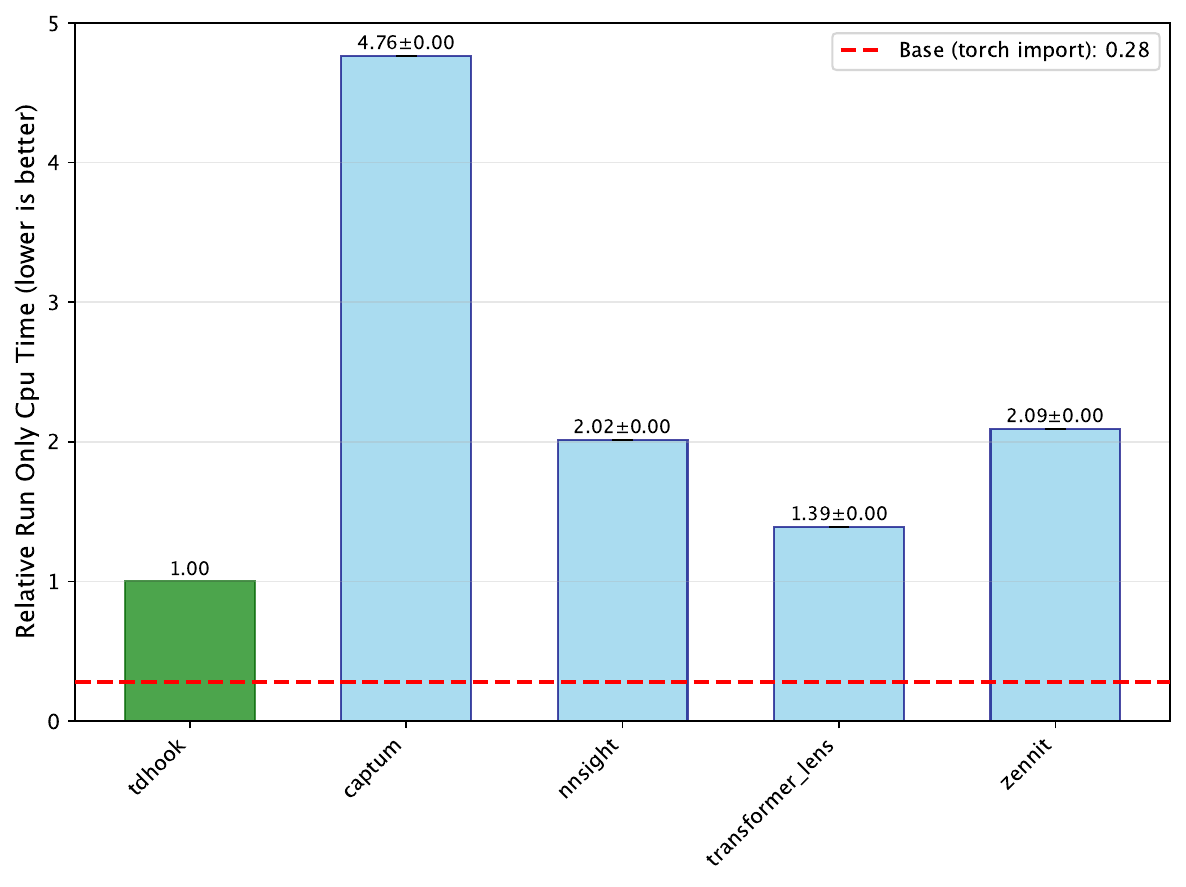}
        \label{fig:benchmarking_run_only_cpu_time}
    }
    \caption{Relative time performance of CPU spawn and run across frameworks.}
    \label{fig:benchmarking_cpu_time_group}
\end{figure}

\begin{figure}[thbp]
    \centering
    \subfloat[GPU spawn time]{
        \includegraphics[width=0.45\textwidth]{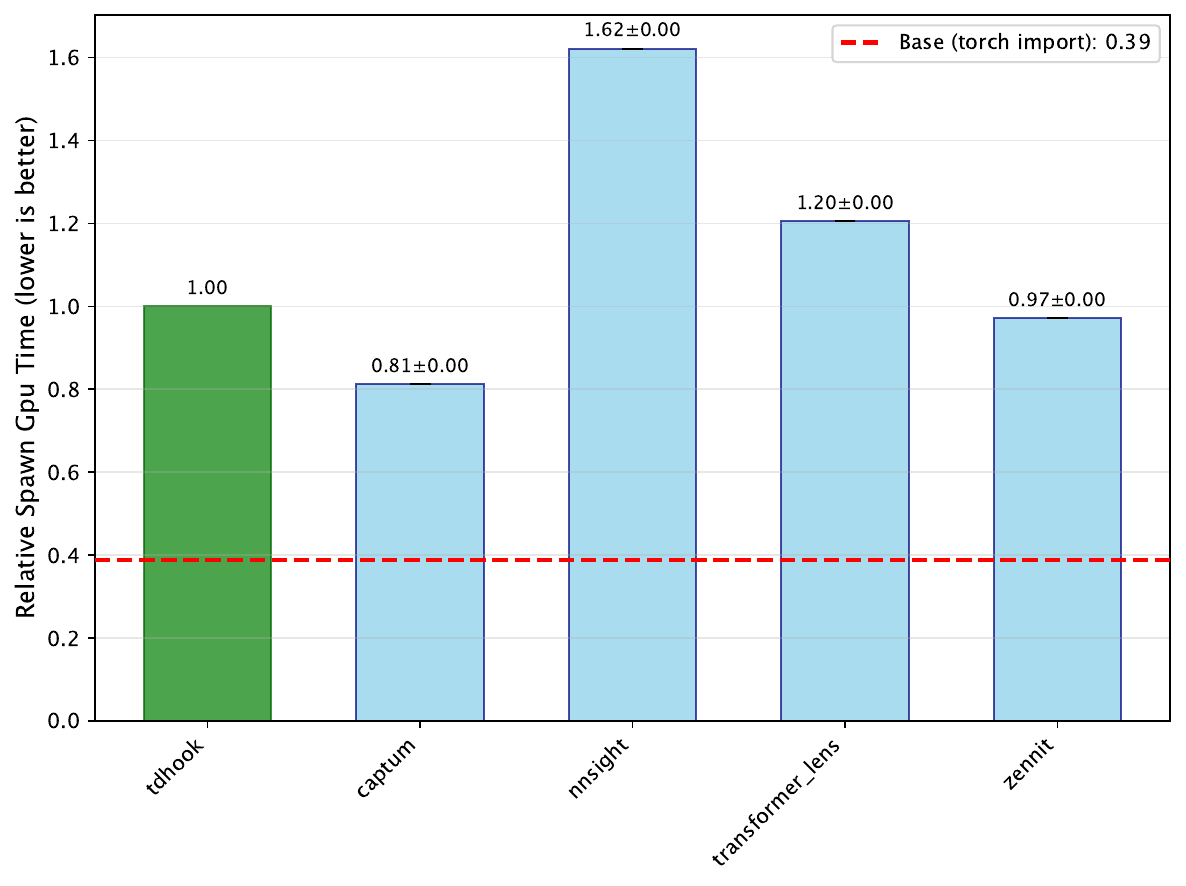}
        \label{fig:benchmarking_spawn_gpu_time}
    }
    \hfill
    \subfloat[GPU run time]{
        \includegraphics[width=0.45\textwidth]{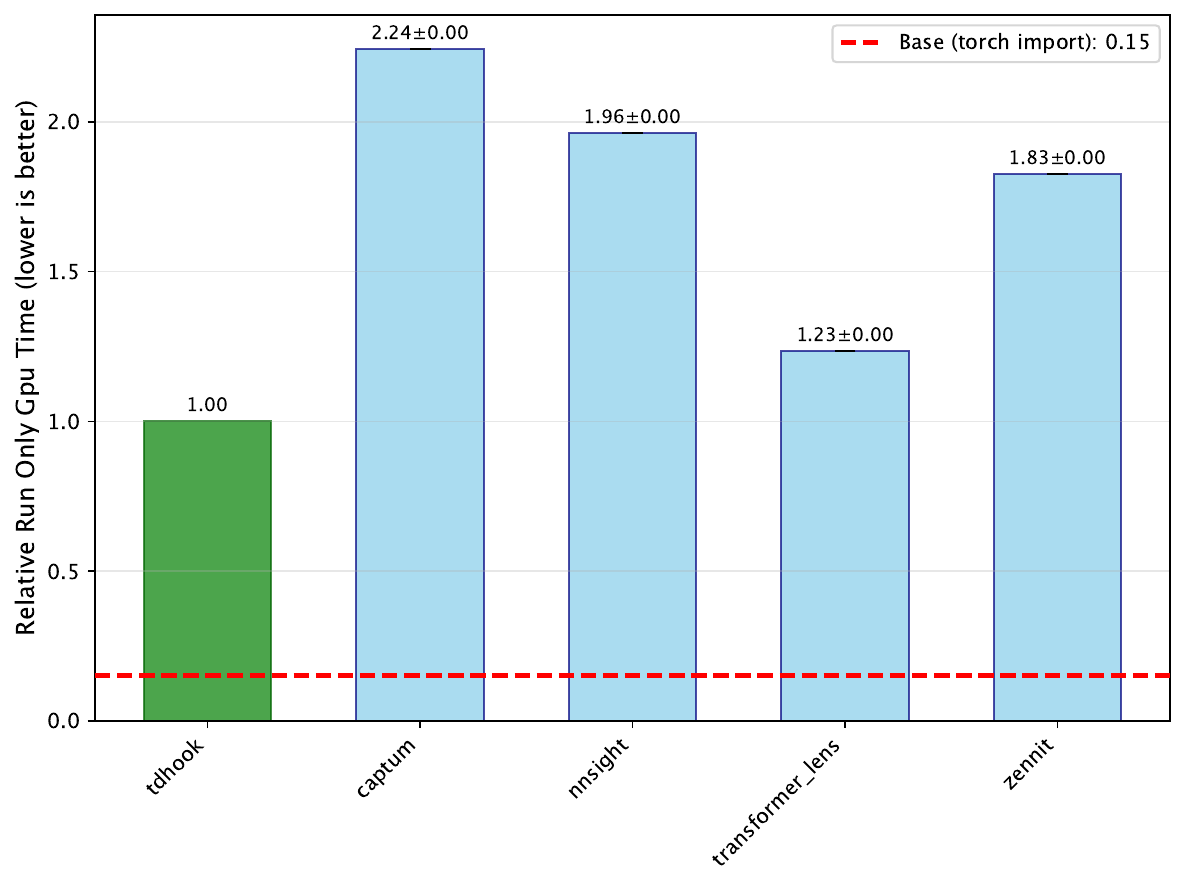}
        \label{fig:benchmarking_run_only_gpu_time}
    }
    \caption{Relative time performance of GPU spawn and run across frameworks.}
    \label{fig:benchmarking_gpu_time_group}
\end{figure}

\begin{figure}[thbp]
    \centering
    \subfloat[CPU spawn RAM]{
        \includegraphics[width=0.45\textwidth]{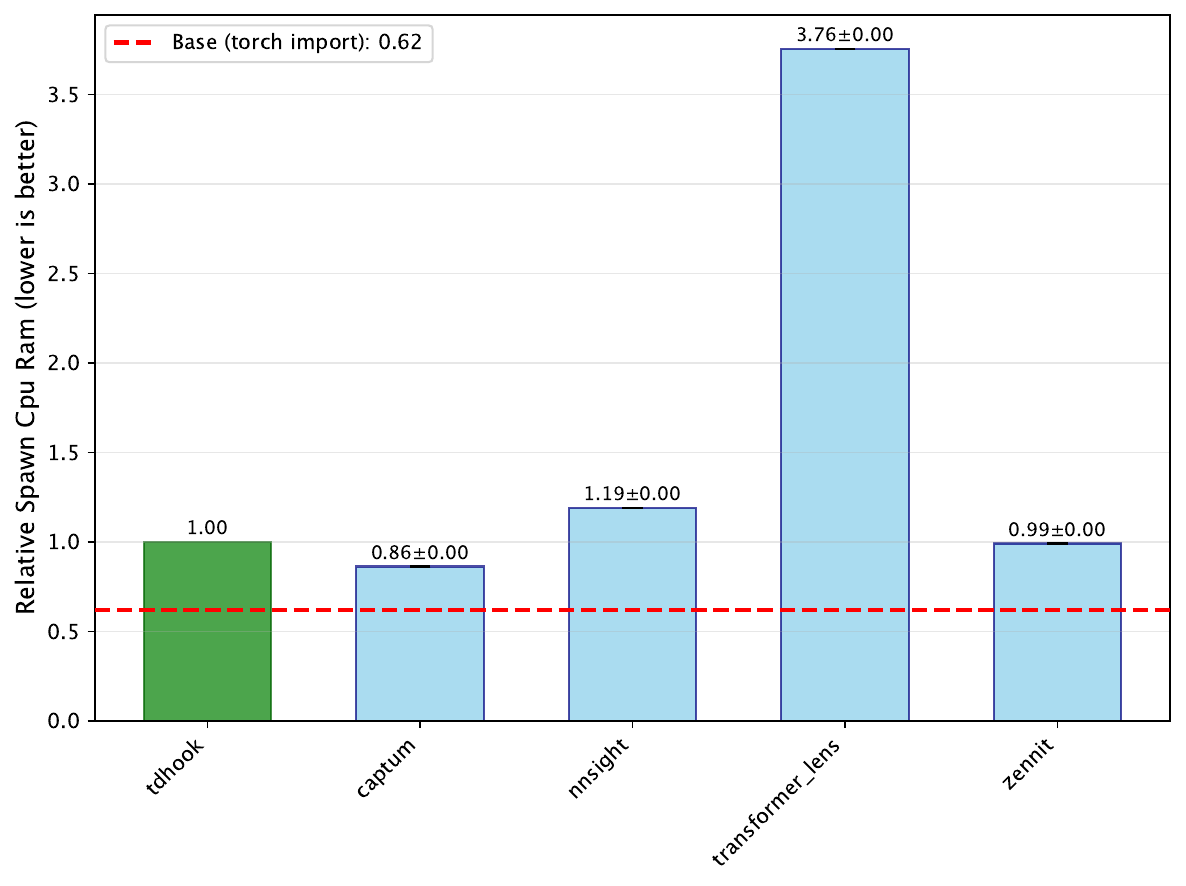}
        \label{fig:benchmarking_spawn_cpu_ram}
    }
    \hfill
    \subfloat[CPU run RAM]{
        \includegraphics[width=0.45\textwidth]{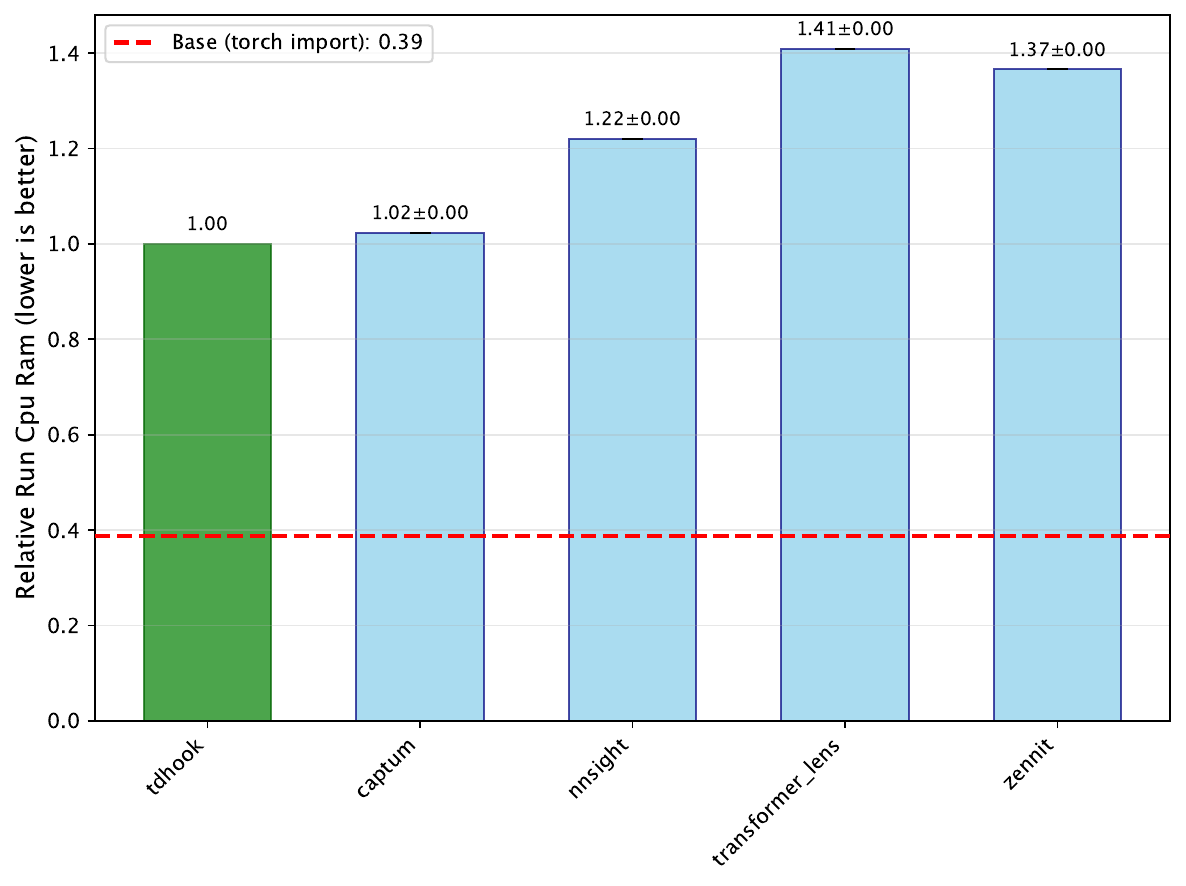}
        \label{fig:benchmarking_cpu_ram}
    }
    \caption{Relative RAM usage of CPU spawn and run across frameworks.}
    \label{fig:benchmarking_cpu_ram_group}
\end{figure}

\begin{figure}[thbp]
    \centering
    \subfloat[GPU spawn RAM]{
        \includegraphics[width=0.45\textwidth]{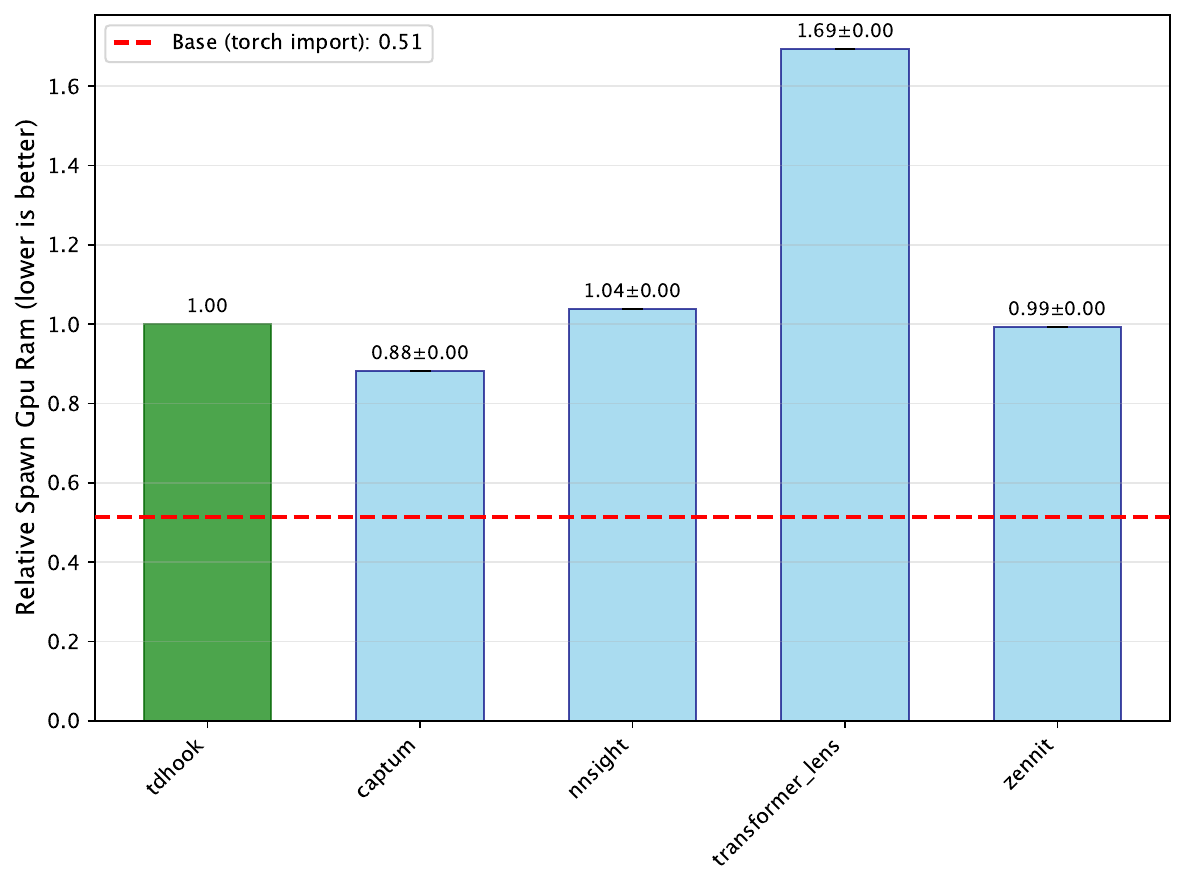}
        \label{fig:benchmarking_spawn_gpu_ram}
    }
    \hfill
    \subfloat[GPU run RAM]{
        \includegraphics[width=0.45\textwidth]{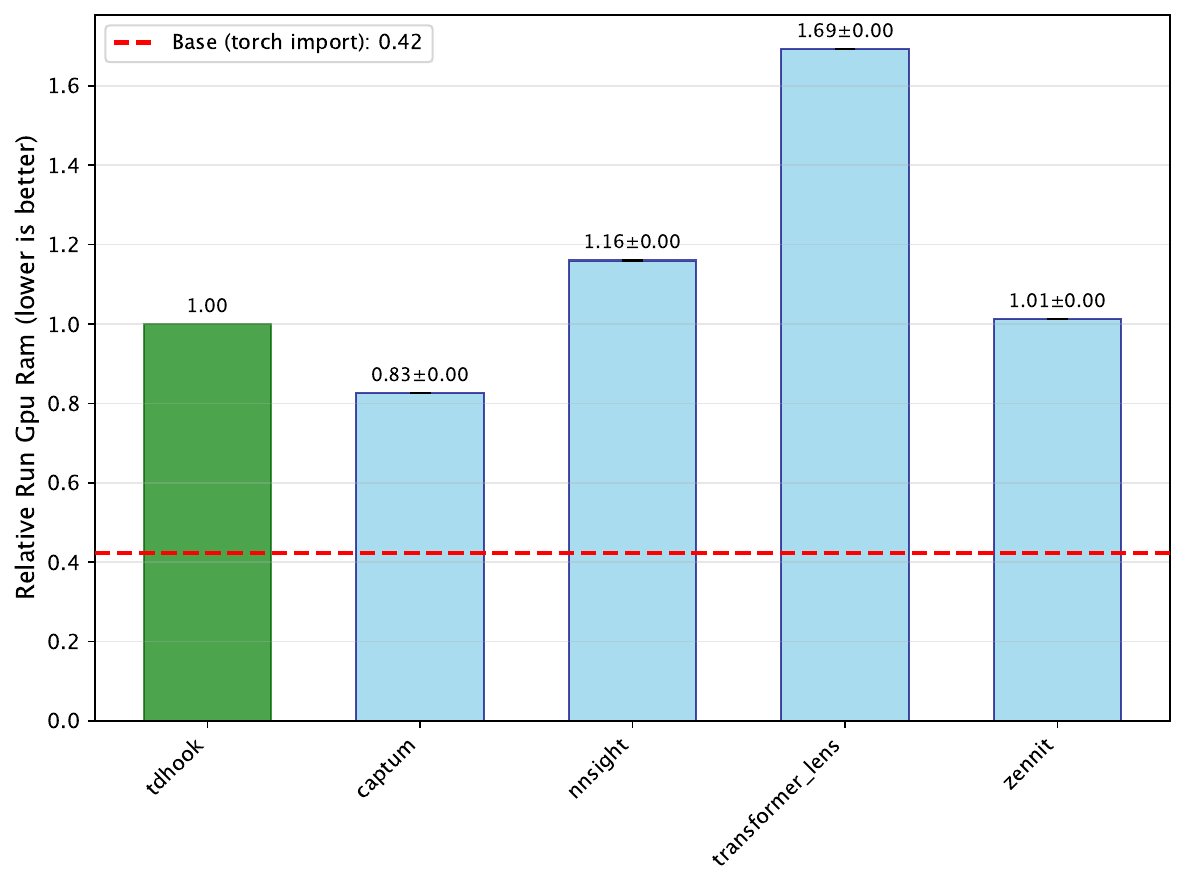}
        \label{fig:benchmarking_run_gpu_ram}
    }
    \caption{Relative RAM usage of GPU spawn and run across frameworks.}
    \label{fig:benchmarking_gpu_ram_group}
\end{figure}

\begin{figure}[thbp]
    \centering
    \subfloat[GPU spawn VRAM]{
        \includegraphics[width=0.45\textwidth]{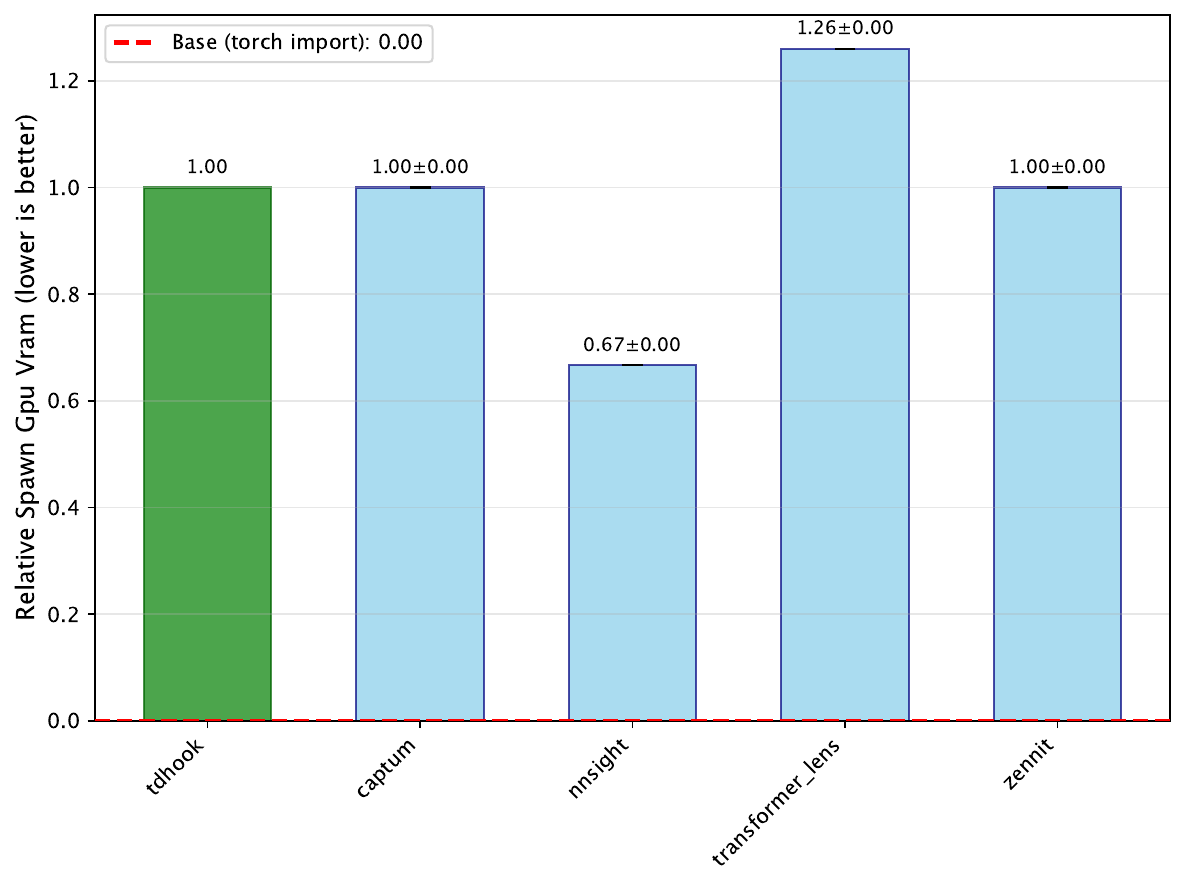}
        \label{fig:benchmarking_spawn_gpu_vram}
    }
    \hfill
    \subfloat[GPU run VRAM]{
        \includegraphics[width=0.45\textwidth]{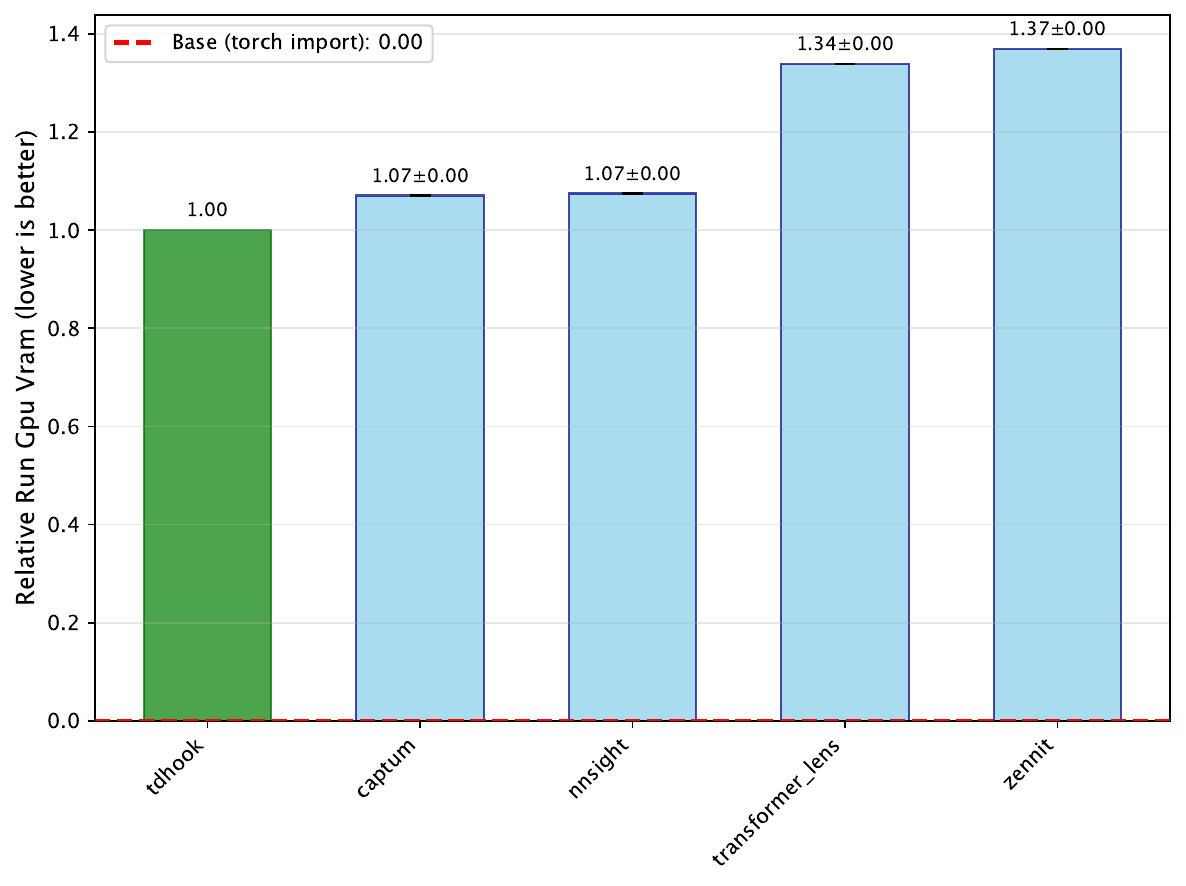}
        \label{fig:benchmarking_run_gpu_vram}
    }
    \caption{Relative VRAM usage of GPU spawn and run across frameworks.}
    \label{fig:benchmarking_gpu_vram_group}
\end{figure}

The performance benchmarking, summarised in Figure \ref{fig:performance_benchmark}, consists of different tasks, each fitting its corresponding library, and all are available in the supplementary material (folder \texttt{scripts/bench/tasks}).
We consider integrated gradients for \texttt{captum}, EpsilonPlus LRP rules for \texttt{zennit}, causal intervention for \texttt{nnsight} and activation caching for \texttt{transformer\_lens}. Each task is run in isolation, meaning that a new process is created for each script, and no parallelisation is used. We launch our performance benchmark on an HPC cluster using SLURM with the following resources: 1 GPU A100 MIG 40GB, 7 CPU Intel Xeon Gold 6330 and 100GB of RAM. Relative performance details can be seen in Figure \ref{fig:benchmarking_cpu_time_group} and Figure \ref{fig:benchmarking_gpu_time_group} for the time metric, in Figure \ref{fig:benchmarking_cpu_ram_group} and Figure \ref{fig:benchmarking_gpu_ram_group} for the RAM metric, and in Figure \ref{fig:benchmarking_gpu_vram_group} for the VRAM metric.

\paragraph{Bundle size benchmark}
The bundle size benchmark, whose results are shown in Figure \ref{fig:bundle_size_benchmark}, measures the additional storage overhead of each framework. For each measurement, we install the specified package in isolation in a Python virtual environment managed by \texttt{uv} \citep{uv}. We then compute the memory and inodes occupied by the virtual environment and compare these measurements to having only \texttt{torch} installed.

\section{Use Cases Details}
\label{app:use_cases_details}

This appendix presents the models used in the use cases presented in Section \ref{sec:use_cases}, and other additional details.

\begin{figure}[thbp]
    \centering
    \includegraphics[width=0.8\textwidth]{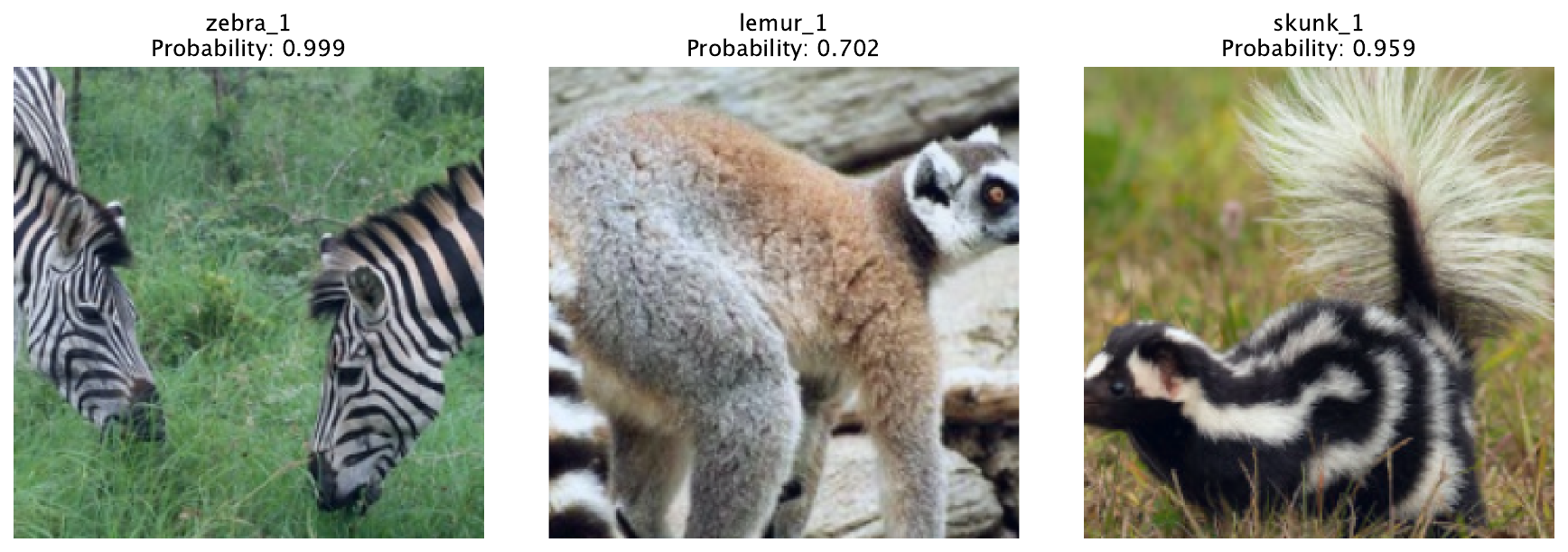}
    \caption{Logistic regression probe evaluation on the animal images.}
    \label{fig:probe_logistic_regression_l1}
\end{figure}

\begin{figure}[thbp]
    \centering
    \includegraphics[width=0.8\textwidth]{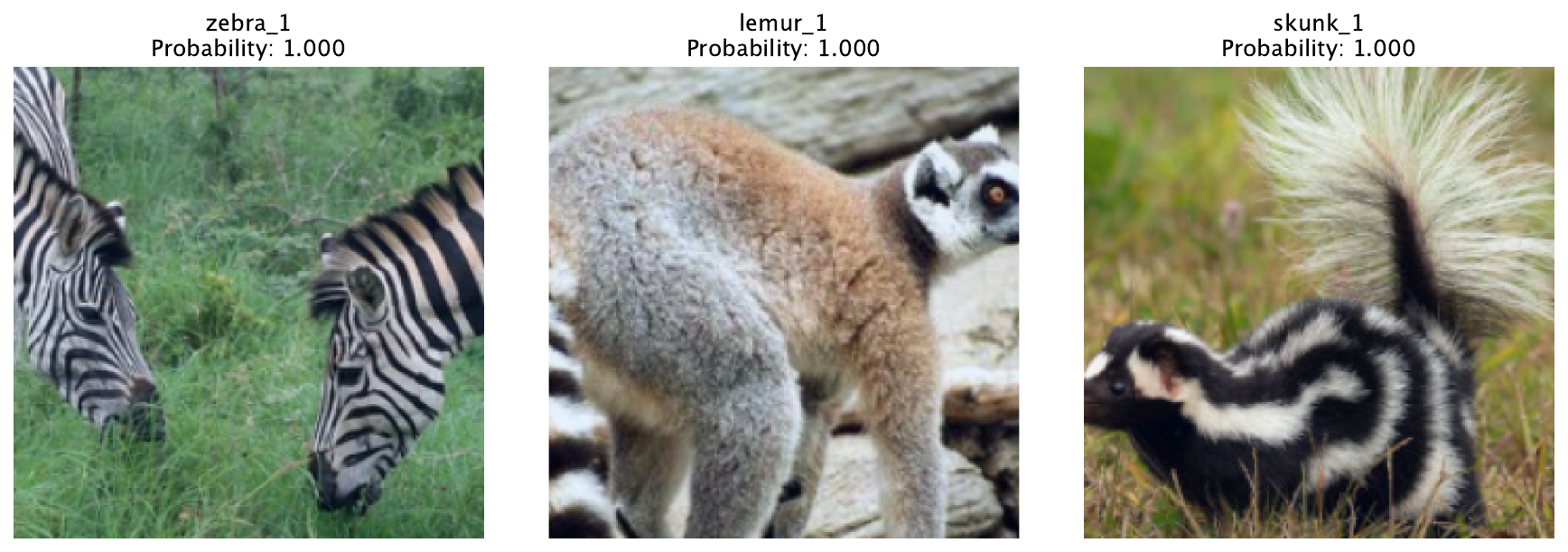}
    \caption{Mean difference probe evaluation on the animal images.}
    \label{fig:probe_mean_difference}
\end{figure}

\paragraph{Concept attribution}
For our analysis, we choose the VGG16 model \citep{Simonyan2014VeryDC}, loaded using the \texttt{timm} library \citep{Wightman_PyTorch_Image_Models}. We train linear probes for concepts on the texture dataset from \citep{Cimpoi2013DescribingTI}, using the \texttt{scikit-learn} library \citep{Pedregosa2011ScikitlearnML}. As pointed out in \citep{Dreyer2023FromHT}, the difference of means (or signal CAV \citep{Pahde2022NavigatingNS}) is more effective to find latent concept directions, as shown in figures \ref{fig:probe_logistic_regression_l1} and \ref{fig:probe_mean_difference}.

\begin{figure}[thbp]
    \centering
    \includegraphics[width=0.8\textwidth]{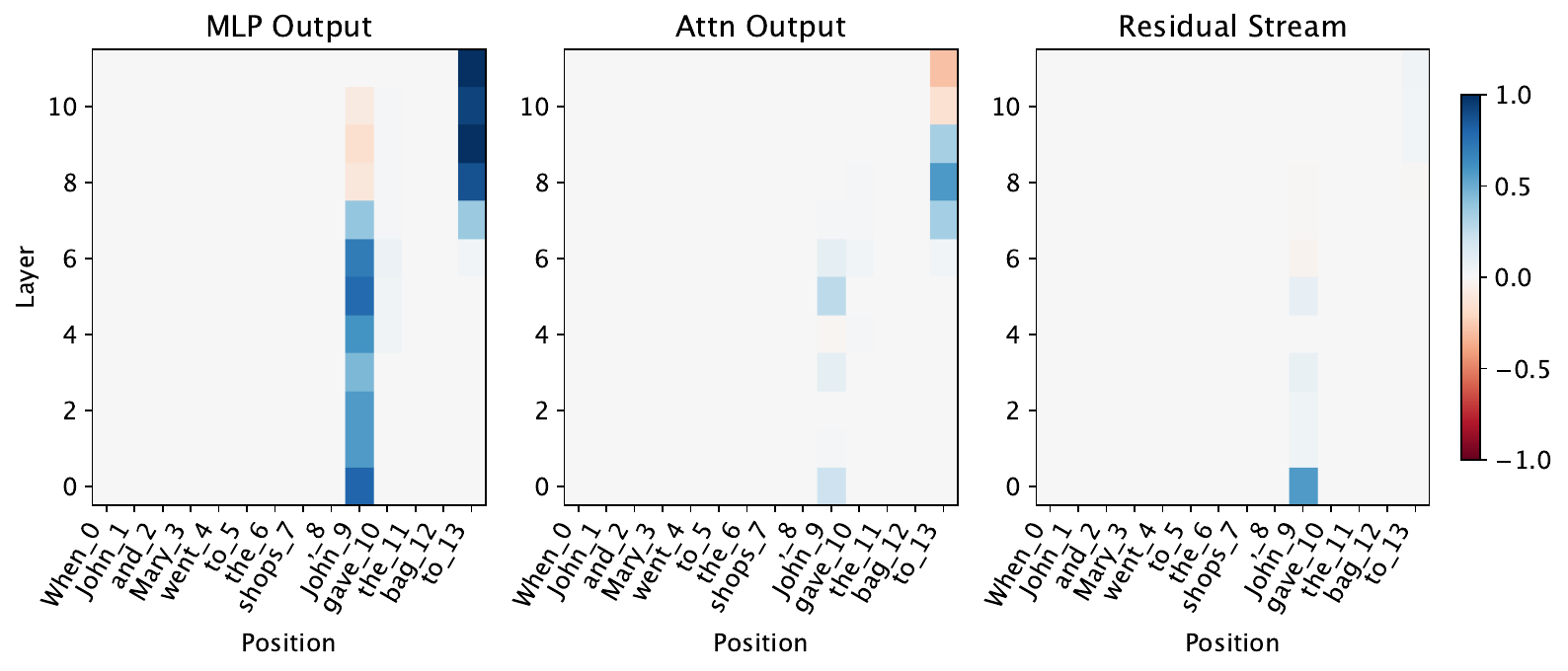}
    \caption{Attribution patching with the RelP method.}
    \label{fig:attribution_patching_relp}
\end{figure}

\begin{figure}[tbhp]
    \centering
    \includegraphics[width=0.8\textwidth]{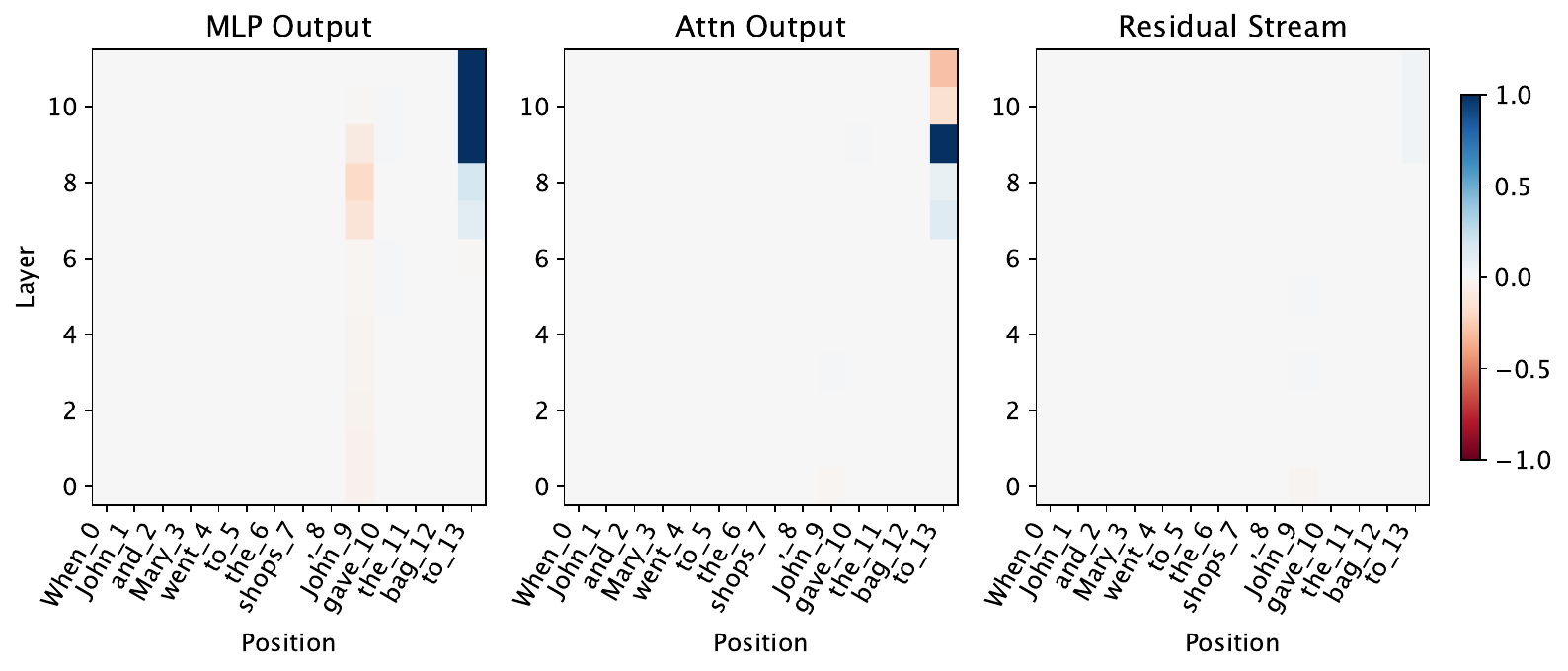}
    \caption{Attribution patching with the RelP method using the AH-rule.}
    \label{fig:attribution_patching_relp_ah}
\end{figure}

\paragraph{Attribution patching}
We run our experiments using the GPT-2 model \citep{Radford2019LanguageMA}, loaded using the \texttt{transformers} library \citep{Wolf2020TransformersSO}.
In order to compute Pearson correlation coefficients, we use the \texttt{scipy} library \citep{2020SciPy-NMeth}.
We show the difference when using the AH-rule instead of the RelP method in figures \ref{fig:attribution_patching_relp} and \ref{fig:attribution_patching_relp_ah}.

\paragraph{Multi-output model}
For studying a multi-output model, we choose a chess model trained with the AlphaZero algorithm \citep{Silver2018AGR}.
While the original proposed algorithm was closed-source, thanks to the Leela team, reproductions were made open source in the lc0 engine \citep{The_LCZero_Authors_LeelaChessZero}. Further initiatives, like Maia networks \citep{McIlroyYoung2020AligningSA}, include the development of human-like chess-playing agents. For our experiments, we choose the \texttt{maia-1900} network, which has two prediction heads, the policy head and the win-draw-lose head.
These heads share a Squeeze-and-Excitation backbone \citep{Hu2017SqueezeandExcitationN}, based on ResNet \citep{He2015DeepRL}, made of 6 blocks.

\paragraph{Composed model}
As we choose to train our agent with the PPO algorithm \citep{Schulman2017ProximalPO}, using the torchrl library \citep{Bou2023TorchRLAD}, we require two networks: the policy and the value.
These two networks are each a multi-layer perceptron with 6 hidden layers of 32 units each, activated by the hyperbolic tangent function.
The model is trained on the inverted double pendulum environment \citep{Todorov2012MuJoCoAP}, from the gymnasium library \citep{Towers2024GymnasiumAS}, derived from cartpole \citep{Barto1983NeuronlikeAE}.
The agent is fully trained, with 1 million frames, before being studied.

\section{Simple Use Cases}
\label{app:simple_use_cases}
This appendix showcases simple, self-contained examples of using TDHook for common interpretability tasks.

\begin{figure}[h]
    \begin{subfigure}{0.26\textwidth}
        \centering
        \includegraphics[width=\linewidth]{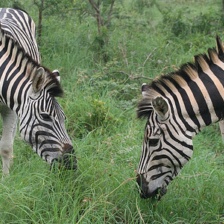}
        \caption{Original image}
        \label{fig:zebra_original}
    \end{subfigure}
    \hfill
    \centering
    \begin{subfigure}{0.3\textwidth}
        \centering
        \includegraphics[width=\linewidth]{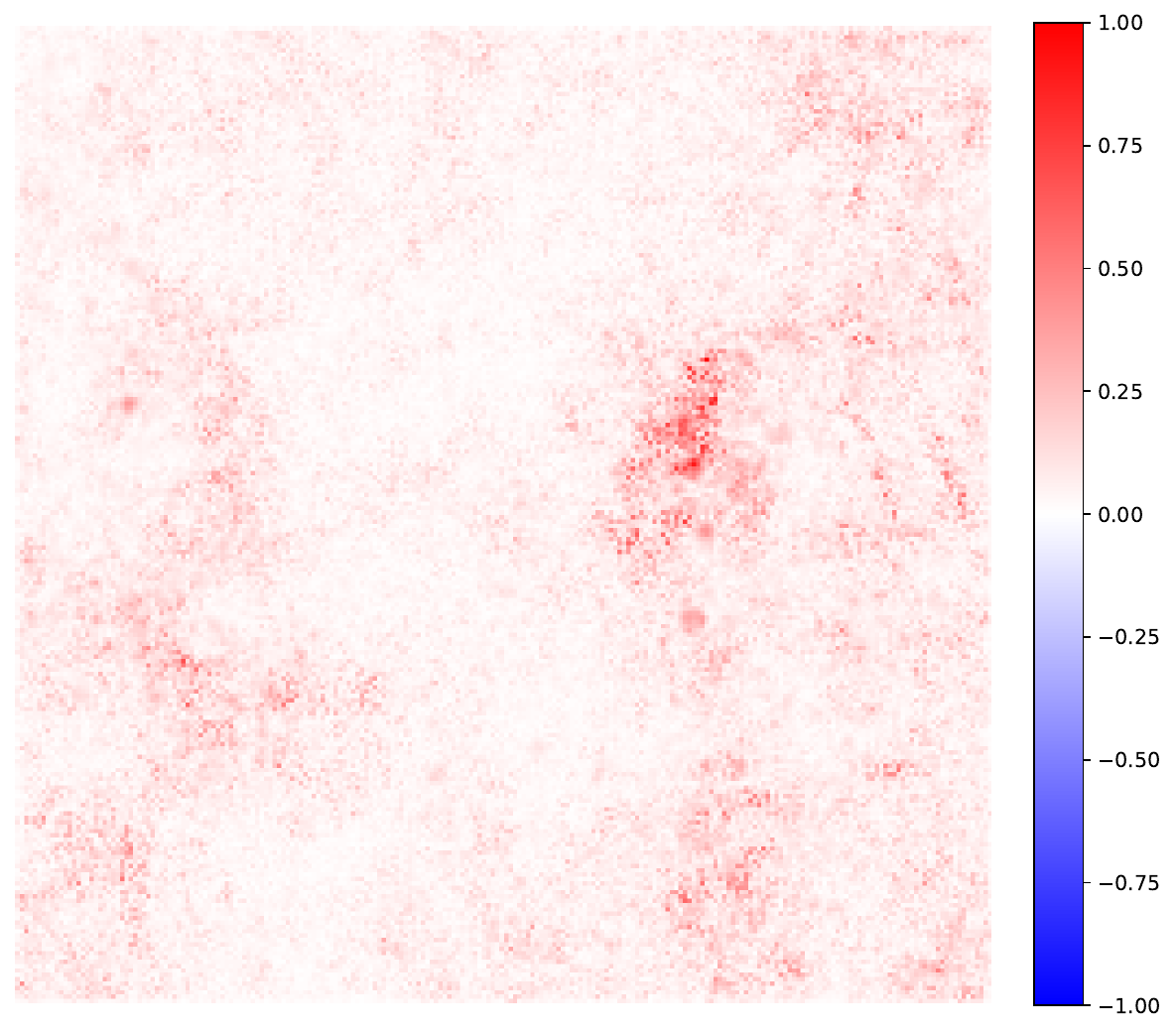}
        \caption{Saliency}
        \label{fig:zebra_saliency}
    \end{subfigure}
    \hfill
    \begin{subfigure}{0.3\textwidth}
        \centering
        \includegraphics[width=\linewidth]{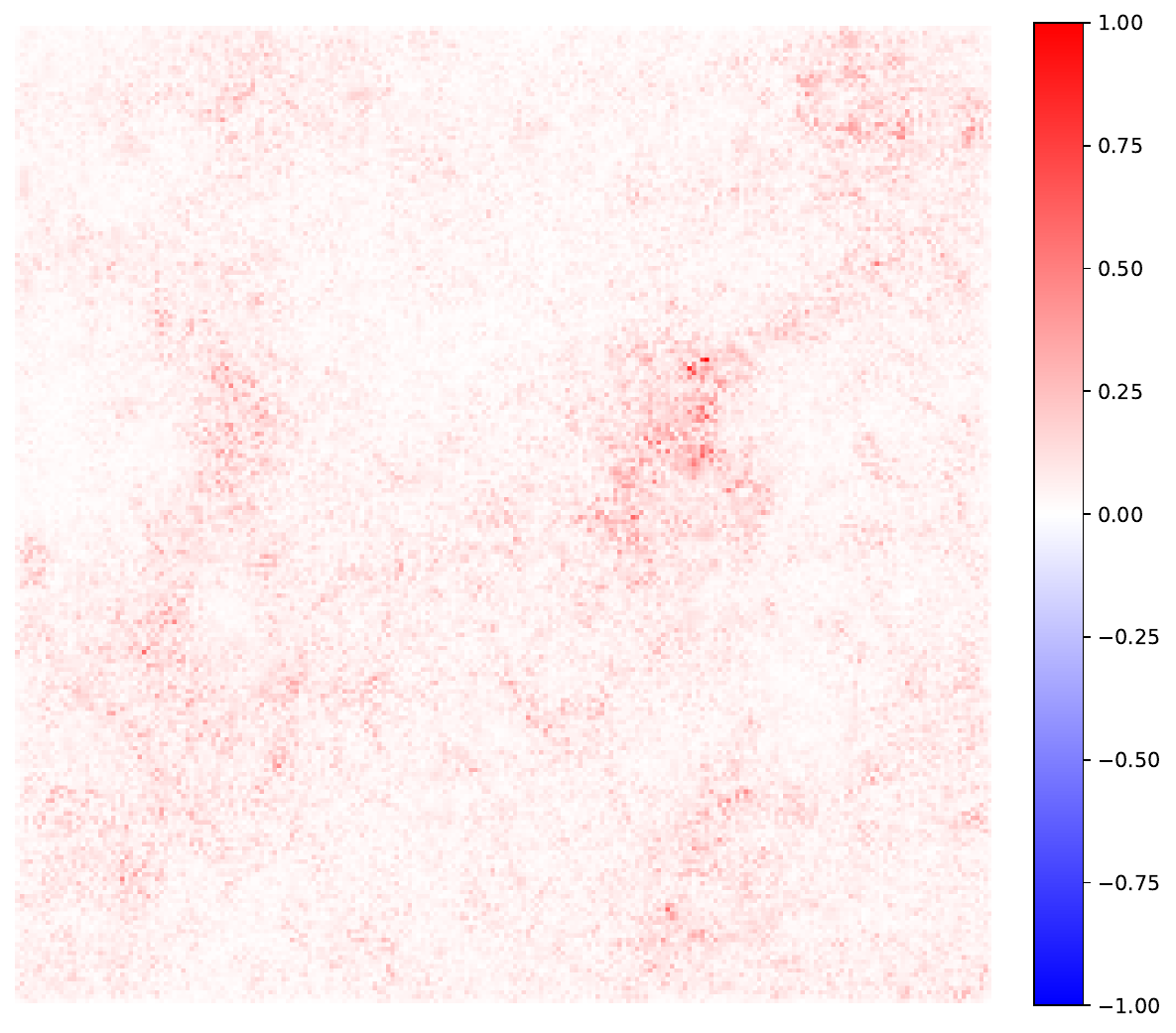}
        \caption{Integrated Gradients}
        \label{fig:zebra_ig}
    \end{subfigure}
    \caption{Comparison of attribution maps for the "zebra" class (index 340) using VGG16. Saliency (left) and Integrated Gradients (right) are computed using the code above.}
    \label{fig:zebra_attr}
\end{figure}

\paragraph{Attribution}
TDHook provides a unified API for attribution methods, making it simple to switch between different techniques. 
In Listing \ref{code:attribution_example}, we demonstrate how to compute attribution maps for a pre-trained VGG16 model.
We highlight the minimal changes required to switch from \texttt{Saliency} (red lines) to \texttt{IntegratedGradients} (green lines), and present both results in Figure \ref{fig:zebra_attr}.
\begin{listing}[H]
\begin{lstlisting}[language=Python]
import torch
import timm
from PIL import Image
from tensordict import TensorDict
|\redbg{from tdhook.attribution import Saliency}|
|\greenbg{from tdhook.attribution import IntegratedGradients}|

# Load model and prepare image
model = timm.create_model("vgg16.tv_in1k", pretrained=True)
data_config = timm.data.resolve_model_data_config(model)
transforms = timm.data.create_transform(**data_config, is_training=False)

image = Image.open("results/simple/zebra_1.jpg").convert("RGB")
image_tensor = transforms(image)

# Define attribution target (zebra class = 340)
def init_attr_targets(targets, _):
    zebra_logit = targets["output"][..., 340]
    return TensorDict(out=zebra_logit, batch_size=targets.batch_size)

# Compute attribution
|\redbg{with Saliency(}|
|\greenbg{with IntegratedGradients(}|
    init_attr_targets=init_attr_targets
).prepare(model) as hooked_model:
    td = TensorDict(
        {
            "input": image_tensor.unsqueeze(0), 
|\greenbg{\hspace{5em}("baseline", "input"): torch.zeros\_like(image\_tensor).unsqueeze(0)}|
        },
        batch_size=1,
    )
    td = hooked_model(td) # Access attribution with td.get(("attr", "input"))
\end{lstlisting}
\caption{Code example showing the switch from \texttt{Saliency} to \texttt{IntegratedGradients}.}
\label{code:attribution_example}
\end{listing}

\paragraph{Steering Vectors}
TDHook supports intervention techniques such as steering vectors which have proven effective for language models \citep{rimsky2023steering}. 
In Listing \ref{code:steering_example}, we demonstrate how to extract a steering vector from GPT-2 that represents the concept of "wealth" and use it to steer the model's generation.
When completing the prompt "I work as a", the model is steered from "writer" to "pilot", reflecting the injected concept\footnote{Additional samples could be used to extract more robust and representative steering vectors.}.

\begin{listing}[H]
\begin{lstlisting}[language=Python]
from transformers import AutoTokenizer, AutoModelForCausalLM
from tensordict import TensorDict
from tdhook.latent import ActivationAddition, SteeringVectors

# Load model and tokenizer
model = AutoModelForCausalLM.from_pretrained("gpt2")
tokenizer = AutoTokenizer.from_pretrained("gpt2")

# Prepare inputs
positive_inputs = tokenizer.encode("I am rich.", return_tensors="pt")
negative_inputs = tokenizer.encode("I am poor.", return_tensors="pt")
base_inputs = tokenizer.encode("I work as a", return_tensors="pt")

# Extract steering vector (rich - poor)
with ActivationAddition(["transformer.h.7.mlp"]).prepare(model) as hooked_model:
    td = TensorDict(
        {
            ("positive", "input"): positive_inputs, 
            ("negative", "input"): negative_inputs
        }, 
        batch_size=1
    )
    td = hooked_model(td)

steering_vector = td.get(("steer", "transformer.h.7.mlp")).sum(dim=0)


# Define steering function
def steer_fn(module_key, output):
    return output + 4 * steering_vector


# Apply steering during inference
with SteeringVectors(["transformer.h.7.mlp"], steer_fn=steer_fn).prepare(model) as hooked_model:
    td = TensorDict({"input": base_inputs}, batch_size=1)
    td = hooked_model(td)

# Compare results
steered_token = td.get(("output", "logits")).max(dim=-1).indices[0, -1]
original_token = model(base_inputs)["logits"].max(dim=-1).indices[0, -1]

print(f"Steered: {tokenizer.decode(steered_token)}")  # Output: "pilot"
print(f"Original: {tokenizer.decode(original_token)}")  # Output: "writer"
\end{lstlisting}
\caption{Code example showing the extraction and application of a steering vector in GPT-2.}
\label{code:steering_example}
\end{listing}

\section{Codebase Outline}
\label{app:code}

This appendix provides an overview of the codebase distributed with the supplementary material.

\paragraph{Source code}
The source code resides in the \texttt{src/tdhook} directory, while the test suite mirrors the same structure under \texttt{tests}.  
The package is organised around three core modules:
\begin{itemize}
  \item \texttt{module.py} defines the \texttt{HookedModel} wrapper among others, which is the core class that wraps a PyTorch model and provides the get-set API or the run context via the \texttt{HookedModuleRun} class.
  \item \texttt{hooks.py} gathers utility functions to create and manipulate hooks, as well as proxy classes to provide cache flexibility.
  \item \texttt{contexts.py} groups context managers that serve as building blocks for the different methods; they enable the creation and management of hooks and modules.
\end{itemize}
Higher-level methods are grouped in dedicated sub-packages: \texttt{attribution} for attribution methods, \texttt{latent} for latent manipulation methods, and \texttt{weights} for weights-based methods.

\paragraph{Main classes interactions}

Each ready-to-use method inherits from the \texttt{HookingContextFactory} class, which is stateless, and is responsible for preparing the initial model by rewriting its forward pass when necessary, composing it with other \texttt{TensorDictModule} instances and registering the necessary hooks.
The \texttt{HookingContextFactory.prepare} method creates a \texttt{HookingContext} instance that spawns a bound \texttt{HookedModel} inside the context and ensures its proper restoration after the context is exited.
The \texttt{HookedModel} class provides a powerful API on its own to manage hooks and cache directly,
and is thus a great point of entry to customise any method. The \texttt{HookedModel.run} method creates a \texttt{HookedModuleRun} instance, which enables the definition of transparent intervention schemes with automatic module execution, hook (de)registration and cache management.

\paragraph{Ready-to-use methods}

Table \ref{tab:methods_summary} provides a summary of all implemented methods organised by category.

\begin{table}[h]
\centering
\caption{Summary of implemented methods by category}
\label{tab:methods_summary}
\resizebox{\textwidth}{!}{%
\begin{tabular}{lcp{0.6\textwidth}}
\toprule
\textbf{Category} & \textbf{Count} & \textbf{Methods} \\
\midrule
Attribution & 13 & \citet{Simonyan2013DeepIC} \citet{Shrikumar2016NotJA} \citet{Sundararajan2017AxiomaticAF} \citet{Dhamdhere2018HowII} \citet{Selvaraju2016GradCAMVE} \citet{Bach2015OnPE} \citet{Lapuschkin2019UnmaskingCH} \citet{Montavon2019LayerWiseRP} \citet{Andol2021LearningDI} \citet{Montavon2015ExplainingNC} \citet{Achtibat2022FromAM} \citet{Mahendran2015VisualizingDC} \citet{Springenberg2014StrivingFS} \\
\midrule
Latent Manipulation & 8 & \citet{Chen2020ConceptWF} \citet{Abnar2020QuantifyingAF} \citet{alain2018understanding} \citet{kim2018interpretability} \citet{Vig2020InvestigatingGB} \citet{belrose2023leace} \citet{Dreyer2023FromHT} \citet{rimsky2023steering} \\
\midrule
Weights & 6 & \citet{Meng2022LocatingAE} \citet{Cunningham2023SparseAF} \citet{Dunefsky2024TranscodersFI} \citet{Ilharco2022EditingMW} \citet{Yeom2019PruningBE} \citet{Pochinkov2024DissectingLM} \\
\bottomrule
\end{tabular}%
}
\end{table}

Attribution methods are implemented in the \texttt{attribution} package:
\begin{itemize}
  \item \texttt{gradient\_attribution/saliency.py}: gradient attribution \citep{Simonyan2013DeepIC} and its gradient-times-input variation \citep{Shrikumar2016NotJA}.
  \item \texttt{gradient\_attribution/integrated\_gradients.py}: integrated gradients \citep{Sundararajan2017AxiomaticAF} and its conditional variant \citep{Dhamdhere2018HowII}.
  \item \texttt{gradient\_attribution/grad\_cam.py}: grad-CAM \citep{Selvaraju2016GradCAMVE}.
  \item \texttt{gradient\_attribution/lrp}: different LRP rules such as LRP-0, LRP-$\epsilon$ z-plus \citep{Bach2015OnPE}, flat \citep{Lapuschkin2019UnmaskingCH}, gamma \citep{Montavon2019LayerWiseRP,Andol2021LearningDI}, w-square \citep{Montavon2015ExplainingNC} and its conditional variant \citep{Achtibat2022FromAM}.
  \item \texttt{activation\_maximisation.py}: activation maximisation \citep{Mahendran2015VisualizingDC}.
  \item \texttt{guided\_backpropagation.py}: guided backpropagation \citep{Springenberg2014StrivingFS}.
\end{itemize}

Latent manipulation methods are implemented in the \texttt{latent} package:
\begin{itemize}
  \item \texttt{activation\_caching.py}: maximally activating samples \citep{Chen2020ConceptWF} and attention visualisation \citep{Abnar2020QuantifyingAF}.
  \item \texttt{probing.py}: linear probing \citep{alain2018understanding} and concept activation vectors \citep{kim2018interpretability}.
  \item \texttt{activation\_patching.py}: causal mediation analysis \citep{Vig2020InvestigatingGB} and latent editing \citep{belrose2023leace,Dreyer2023FromHT}.
  \item \texttt{steering\_vectors.py}: steering vectors \citep{rimsky2023steering}.
\end{itemize}

Weights-related methods are implemented in the \texttt{weights} package:
\begin{itemize}
  \item \texttt{adapters.py}: ROME \citep{Meng2022LocatingAE}, sparse autoencoders \citep{Cunningham2023SparseAF} and transcoders \citep{Dunefsky2024TranscodersFI}.
  \item \texttt{task\_vectors.py}: task vectors \citep{Ilharco2022EditingMW}.
  \item \texttt{pruning.py}: relevance-based pruning \citep{Yeom2019PruningBE} and circuit pruning \citep{Pochinkov2024DissectingLM}.
\end{itemize}

Future work could extend attribution with occlusion-based methods \citep{Zeiler2013VisualizingAU}, latent activation with AtP* \citep{Kramar2024AtPAE}, or weights-based methods with crosscoders that generalise adapters to multiple layers and models \citep{lindsey2024sparse}.

\paragraph{Scripts}
Executable scripts are located in the \texttt{scripts} directory, and their outputs, like figures or logs, are stored in the \texttt{results} directory.  
The \texttt{scripts/bench} subdirectory hosts the benchmarking pipeline, with scripts to run the different methods and plot the results. In particular, each task has corresponding scripts, one per library and per task, in the \texttt{scripts/bench/tasks} directory, which define variations of the task, such as the batch size or model hyperparameters.
The bundle size benchmark utilises an empty project \texttt{scripts/bundle\_test} to install each package in isolation. 
The example use-cases used in Section \ref{sec:use_cases} are located in the \texttt{scripts/torchrl} and \texttt{scripts/lczerolens}, respectively.

\section{Software}
\label{app:software}

This appendix summarises the software used in this work.

\paragraph{Library core}
First, in order to build our library, we heavily relied on \texttt{torch}, \citep{Ansel2024PyTorch2F} as a building block for manipulating models and tensors, at the heart of our library are the \texttt{torch} hooks. And as previously mentioned, we focused our efforts around the \texttt{tensordict} library \citep{Bou2023TorchRLAD}, using \texttt{TensorDict} to manipulate the different artifacts and \texttt{TensorDictModule} to implement the different methods.

\paragraph{Library tests}
We ran all our tests using \texttt{pytest} \citep{pytest}.
In addition, we used some of the aforementioned libraries to compare our implementation for different methods. 
LRP rules were compared against \texttt{zennit} implementations \citep{Anders2021SoftwareFD}. Attribution methods were compared against \texttt{captum} implementations \citep{Kokhlikyan2020CaptumAU} and the get-set API was compared against \texttt{nnsight} \citep{FiottoKaufman2024NNsightAN} using different intervention schemes.

\paragraph{Scripts}
For demonstrating use cases of the library, we used \texttt{timm} \citep{Wightman_PyTorch_Image_Models} to load vision models, \texttt{transformers} \citep{Wolf2020TransformersSO} to load language models, \texttt{torchrl} \citep{Bou2023TorchRLAD} to train a simple RL agent using the PPO algorithm and \texttt{lczerolens} \citep{Yoann_Poupart_LCZeroLens} to manipulate the chess networks. The linear probes were trained using the linear regression model from the \texttt{scikit-learn} library \citep{Pedregosa2011ScikitlearnML}. In order to compute Pearson correlation coefficients, we used the \texttt{scipy} library \citep{2020SciPy-NMeth}.
For plotting our benchmarking and study results, we used \texttt{matplotlib} \citep{Hunter2007MatplotlibA2} and \texttt{seaborn} \citep{Waskom2021SeabornSD}. 
For manipulating the experiment results, we used \texttt{pandas} \citep{the_pandas_development_team_2025_15831829} and \texttt{numpy} \citep{Harris2020ArrayPW}.
We used \texttt{uv} \citep{uv} to run our scripts and manage our dependencies.

\end{document}